\documentclass[runningheads]{llncs}

\usepackage{eccv}

\usepackage{eccvabbrv}

\usepackage{graphicx}
\usepackage{booktabs}

\usepackage[accsupp]{axessibility}  %

\usepackage{hyperref}

\usepackage{orcidlink}

\usepackage{threeparttable}
\usepackage{multicol}
\usepackage{multirow}
\usepackage{colortbl}
\usepackage{longtable}
\usepackage{graphicx}
\usepackage{placeins}
\definecolor{mygreen}{RGB}{93, 141, 74}
\definecolor{myred}{RGB}{146, 86, 95}
\usepackage{subcaption}
\usepackage{floatflt}
\usepackage{wrapfig}
\usepackage{adjustbox}

\begin{document}

\title{Tiny Models are the\\ Computational Saver for Large Models}

\author{Qingyuan Wang\inst{1}\orcidlink{0000-0002-7879-4328} \and Barry Cardiff\inst{1}\orcidlink{0000-0003-1303-8115} \and Antoine Frappé\inst{2}\orcidlink{0000-0002-0977-549X} \and\\ Benoit Larras\inst{2}\orcidlink{0000-0003-2501-8656}\and Deepu John\inst{1}\orcidlink{0000-0002-6139-1100}}

\authorrunning{Q. Wang et al.}

\institute{University College Dublin, Ireland \\ \email{qingyuan.wang@ucdconnect.ie; \{barry.cardiff,deepu.john\}@ucd.ie}\and
    Univ. Lille, CNRS, Centrale Lille, Junia, Univ. Polytechnique Hauts-de-France, UMR 8520-IEMN, France
    \\
    \email{\{antoine.frappe,benoit.larras\}@junia.com}}

\maketitle

\begin{abstract}

This paper introduces TinySaver, an early-exit-like dynamic model compression approach which employs tiny models to substitute large models adaptively. Distinct from traditional compression techniques, dynamic methods like TinySaver can leverage the difficulty differences to allow certain inputs to complete their inference processes early, thereby conserving computational resources. Most existing early exit designs are implemented by attaching additional network branches to the model's backbone. Our study, however, reveals that completely independent tiny models can replace a substantial portion of the larger models' job with minimal impact on performance. Employing them as the first exit can remarkably enhance computational efficiency. By searching and employing the most appropriate tiny model as the computational saver for a given large model, the proposed approaches work as a novel and generic method to model compression. This finding will help the research community in exploring new compression methods to address the escalating computational demands posed by rapidly evolving AI models. Our evaluation of this approach in ImageNet-1k classification demonstrates its potential to reduce the number of compute operations by up to 90\%, with only negligible losses in performance, across various modern vision models.
\keywords{Dynamic Model Architecture \and Model Compression \and Early Exit}
\end{abstract}

\section{Introduction}
\label{sec:intro}

\begin{figure}[t]
\centering
\includegraphics[width=\linewidth]{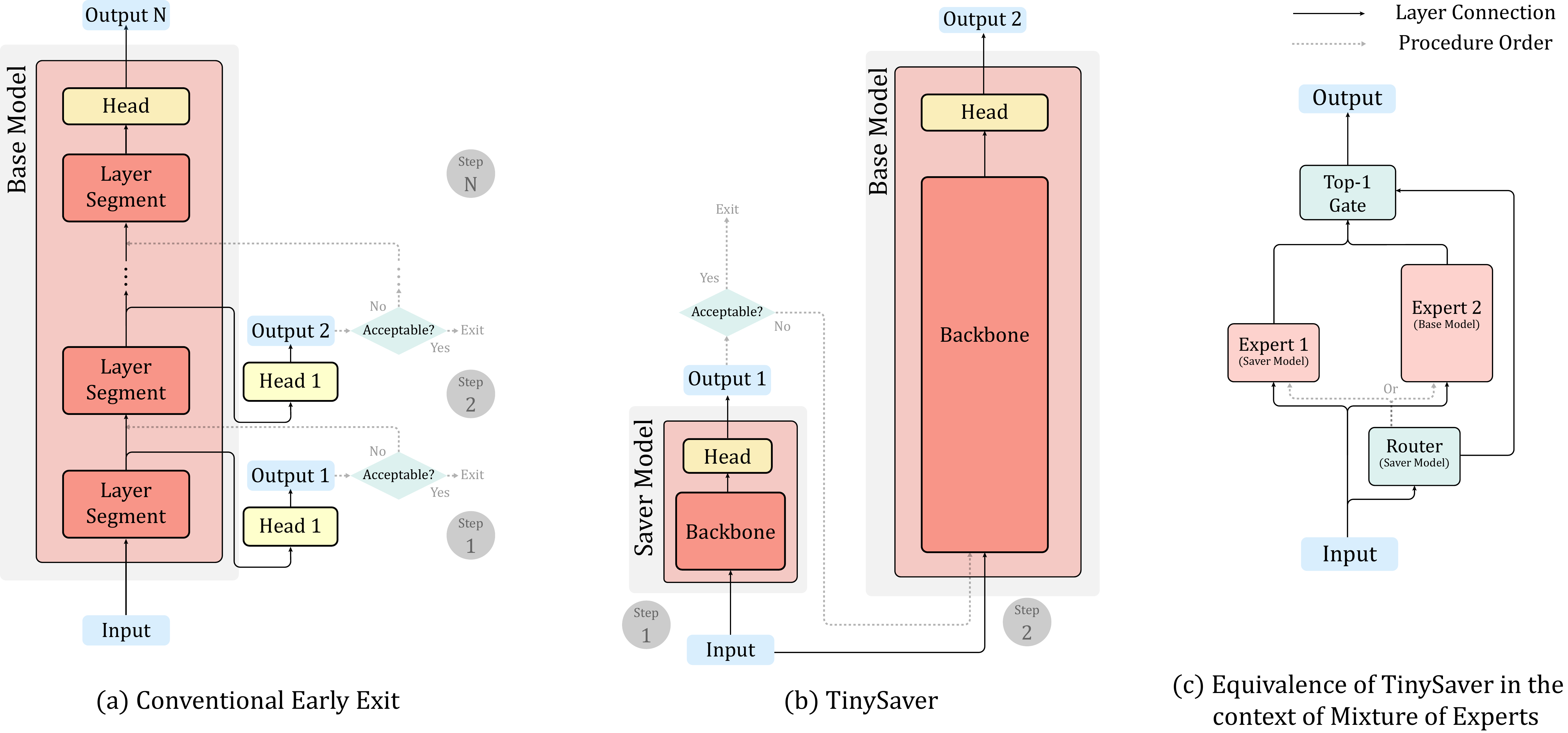}
\caption{Illustration of TinySaver with the comparison to related concept of Early Exit and Mixture of Experts.}
\label{fig:simple-case}

\end{figure}
\footnotetext{Code available at \url{https://github.com/QingyuanWang/tinysaver}}As we marvel at the exceptional performance of large-scale AI models, their rapidly increasing sizes pose significant challenges in terms of computational demands. In 2023, cutting-edge vision models have attained up to 90\% accuracy on the ImageNet-1k dataset, utilizing 362 million parameters\cite{ding_davit_2022}. This parameter count is sixfold that of the ResNet-152 model\cite{he_deep_2016}, which was deemed large, eight years prior. The AI models in the Natural Language Processing (NLP) domain are expanding even more swiftly, now encompassing tens of billions of parameters \cite{openai_gpt-4_2023}. These expansive models necessitate substantial computational resources. Furthermore, the swift escalation in AI services utilizing large models has a notable impact on society's overall energy consumption\cite{yokoyama_investigating_2023}. This situation underscores the urgency to develop more efficient methods for computational reduction.

Early Exit (EE)\cite{teerapittayanon_branchynet_2017,laskaridis_spinn_2020,kaya_shallow-deep_2019,panda_conditional_2016} based methods are emerging as a promising approach for computational reduction, offering savings in a dimension distinct from traditional compression methods. It belongs to the category of dynamic model inference techniques\cite{han_dynamic_2021} that enables input samples to traverse different data paths. This allows for case-by-case treatment of each input, recognizing the inherent variability in sample complexity (e.g., blurred or distorted images present more challenges than clear images). Some samples, being less complex, can be effectively processed by smaller models, while more complex samples necessitate larger models. In contrast to conventional methods that uniformly apply compression across all samples, EE provides the flexibility to tailor compression specifically to each input. As depicted in \cref{fig:simple-case}(a), an EE-enabled system employs multiple exits, each attempting to produce outputs before reaching the final model head. The inference process ceases as soon as the output satisfies predefined exit criteria. Moreover, these early exits can be easily reconfigured, allowing for adjustable compression rates to meet varying needs post-deployment or even in real time. Beyond merely scaling performance, dynamic models offer the potential for integrating multi-task or domain-specific models, sharing a common backbone\cite{ma_modeling_2018}. The primary drawback of EE models is the requirement for additional structural components, therefore it does not reduce model size. However, this is less concerning as a large number of model computations will consume more energy, contribute to pollution and strain electricity grids.

The unique benefits of EE models underscore their considerable potential for various applications. However, its development remains in a nascent stage. Predominantly, existing EE approaches involve integrating auxiliary branches into middle layers of a base model's backbone, with the expectation that these small exits can yield acceptable predictions from intermediate backbone features\cite{scardapane_differentiable_2020,laskaridis_adaptive_2021}. Nonetheless, these exit branches from the initial layers often struggle to perform satisfactorily. This is partly because they rely on a backbone that is typically neither designed nor trained to support such exits. Some studies have attempted to incorporate exit strategies into the backbone design, jointly training exits and the base model\cite{chen_learning_2020,wang_skipbert_2022,liao_global_2021}. However, these approaches are not enough to produce high-performance exits at the level of modern models as joint design will ask shallow backbone layers to provide both high and low-level features at the same time. Additionally, in the vision domain, early-stage features are often limited in their receptive field, lacking the global view and high-level semantic understanding necessary for optimal performance. Consequently, these limitations hinder the advancement and effectiveness of EE based methods.

Moreover, many existing implementation of EEs are application-specific designs\cite{hong_panda_2021,liu_fastbert_2020}. The design and training of these exits are coupled with their main models, posing challenges in applying EE concepts to other models and applications. To facilitate wider employment of  EEs, we rethink existing methodologies and recognize the potential of decoupling this concept from specific model architectures, thereby transforming it into a generic dynamic model compression method.

The core concept of EE is to save computation by allowing sub-networks to handle parts of the task with comparable performance. Similarly, efficient-focused independent models \cite{tan_efficientnet_2019-1,liu_efficientvit_2023,cai_efficientvit_2023,li_rethinking_2023,howard_searching_2019} are designed to process simpler inputs with significantly lower complexity. By implementing an exiting policy that directs only simpler samples to tiny models, we can establish an EE-like system to minimize computation for straightforward inputs. Additionally, this exiting policy is configurable to balance different efficiency-accuracy trade-offs. As a result, combining tiny models with EE principles forms a dynamic, configurable compression method. It can be applied to larger models within the same task, efficiently process simple inputs while reserving complex tasks for the larger model without making too many mistakes. In essence, tiny models can act as computational savers for larger models by successfully undertaking a portion of the workload with equivalent performance.

To achieve this goal, we introduce TinySaver, a system that leverages pre-trained efficient models as a component of the EE framework, aimed at reducing the overall computation in inference. As depicted in \cref{fig:simple-case}(b), the computational process initiates with a tiny saver model, which acts as the primary early exit point. If the result from the saver model meets the acceptance criteria (step 1), the process halts; otherwise, the larger base model will be invoked (step 2). This approach contrasts with prior early-exit methodologies (illustrated in \cref{fig:simple-case}(a)), where a complete efficient model guarantees that overall performance is not inferior. Despite their limited size, these tiny models significantly outperform traditional exit points embedded within the backbone, thereby enhancing overall system performance. Importantly, TinySaver is adaptable to model architectures that are difficult to be segmented. Small exits cannot be attached to shallow parts if the model necessitate the feature from deep levels\cite{ronneberger_u-net_2015,carion_end--end_2020}. However, in our case, the employed complete model can still work.

Moreover, in the context of TinySaver, all models are pre-trained using their original methods, eliminating the need for additional training. It also effectively decouples early exits from specific applications, facilitating the straightforward incorporation of the concept of dynamic model compression into evolving AI models. This low development cost allows easy application to new models, similar to common compression methods\cite{cheng_survey_2020-1}. Additionally, TinySaver's performance can be upgraded with newly proposed efficient models. This approach not only addresses current challenges but also fosters new perspectives for the development of efficient and scalable AI systems.

The primary objective of introducing TinySaver is to overcome performance limitations encountered when implementing the EE concept. However, this method can also be explained as a specialized Mixture of Experts (MoE)\cite{fedus_switch_2022, gururangan_scaling_2023, komatsuzaki_sparse_2023, mustafa_multimodal_2022, shazeer_outrageously_2017} for computation reduction of a given model. In this context, we employ non-uniform pre-trained models as experts and the smaller expert also works as the router as \cref{fig:simple-case}(c). Compared to trained routers which are commonly used in MoE, our routing methods have obvious superiority of performance and system building cost in pure classification tasks and are competitive in other tasks.

We validate our idea by designing dynamic compression systems for vision models. This design demonstrates substantial computational savings for various modern models, achieved with negligible or no loss in performance.

The contribution of this work can be summarized as follows:
\begin{enumerate}
    \item We propose a simple method to integrate pre-trained tiny models with the EE concept, creating a universal method for computational compression with dynamic model architectures.
    \item We analyse the eligibility and efficiency of using tiny models for computational saving purpose in the context of EE and MoE.
    \item We introduce a methodology for selecting the most effective saver model to compress a specified base model. Additionally, we discuss the possible extension of the proposed idea.
    \item Our research demonstrates a significant reduction in computational requirements for contemporary vision models, thereby validating the practicality and efficiency of dynamic model compression.
\end{enumerate}

\section{Related Work}
\label{sec:related-work}

\subsection{Efficient Vision models}

In recent years, the performance of vision models has advanced rapidly, with numerous studies continuously setting new benchmarks\cite{ding_davit_2022,woo_convnext_2023-1,tu_maxvit_2022,liu_swin_2021-1,dosovitskiy_image_2021-1,dai_coatnet_2021,he_deep_2016}. Alongside these developments, another strand of research focuses on efficiency, striving to achieve optimal performance within a framework of acceptable complexity\cite{tan_efficientnet_2019-1,liu_efficientvit_2023,cai_efficientvit_2023,li_rethinking_2023,howard_searching_2019}. However, performance and efficiency often are conflicting and difficult to be covered in the same design. Furthermore, leading models typically rely on well-designed training recipe\cite{he_masked_2022,woo_convnext_2023-1,zhang_mixup_2018-1,yun_cutmix_2019-1,cubuk_autoaugment_2019}, access to large private datasets\cite{sun_revisiting_2017}, and comprehensive hyper-parameter tuning\cite{tan_efficientnet_2019-1,elsken_neural_nodate}. While most architectures are scalable\cite{dehghani_scaling_2023,he_masked_2022,woo_convnext_2023-1}, downstream applications usually have limited version choices, restricted to publicly available pre-trained models, which may not be ideally scaled for specific needs.
\subsection{Dynamic model inference}

Dynamic model inference\cite{laskaridis_adaptive_2021,han_dynamic_2021,xu_survey_2023,wu_blockdrop_2019,wang_skipnet_2018,cai_once-for-all_2020} introduces the concept of a dynamic computing graph, enabling each input sample to be directed to the most suitable path. This approach uniquely facilitates sample differentiation, assigning each to appropriate layers for computational savings, enhanced performance, or multi-tasking.
\subsubsection{Early Exits (EE)}
Early Exits\cite{teerapittayanon_branchynet_2017,laskaridis_spinn_2020,kaya_shallow-deep_2019,panda_conditional_2016,yang_resolution_2020}, as a leading method in dynamic inference for computational saving, propose the termination of predictions as soon as the necessary computations are completed. While promising as a general model compression technique, research in this domain is still developing. Many implementations of EE are employed in dedicated designs to address efficiency\cite{hang_msnet_2023,hong_panda_2021,phuong_distillation-based_2019,xing_early_2020,chen_cf-vit_2022, han_dynamic_2023, wang_not_2021,han_learning_2022}. Some other work explore the feasibility of integrating EE with more  cases\cite{laskaridis_hapi_2020,laskaridis_spinn_2020,kaya_shallow-deep_2019,wang_glance_2020, zhang_basisnet_2021}. However, challenges remain in generalizing early exits, such as determining the optimal exit points and constructing efficient exits. Meanwhile, although not enabled by every work, the routing policy in early exits, is usually adjustable post-training and even during runtime, allowing users to finely tune the system's behavior to meet specific requirements. DyCE\cite{wang_dyce_2024} was introduced to search the best routing configurations, i.e. which exit to use, for different trade-off preferences in real time. The research presented in this paper, however, tackles the latter challenge, focusing on exploring what are most efficient exits.

\subsubsection{Sparsely-Gated Mixture-of-Experts (MoE)}
Sparsely-Gated MoE\cite{fedus_switch_2022, gururangan_scaling_2023, komatsuzaki_sparse_2023, mustafa_multimodal_2022, shazeer_outrageously_2017} is another recent popular dynamic model architecture usually introduced for scaling models to larger sizes. It employs multiple sub-models but only some of them are invoked during the inference according to router's evaluation for a given sample. Therefore the model capacity can be increased without proportionally increasing the computation. Most successful models with MoE are language models, but there are some related work in vision models such as V-MoE\cite{riquelme_scaling_2021,daxberger_mobile_2023} and Swin-MoE\cite{hwang_tutel_2023}. Our work is proposed for saving computation for a pre-trained model which differs from the common MoE. However, it can be considered a special case in the context of MoE where the saver model simultaneously serves as the router and an expert.

\section{Tiny Model as the Computational Saver}
TinySaver is a simple idea. It only requires two models trained for the same task stacked together. We compare this idea with EE and MoE to demonstrate that using a tiny model as is can be 1) a computational saving replacement of large models for many inputs without a large performance compromise. 2) a competitive router to identify reliable predictions, especially for general classification tasks. We also propose a method to find the most appropriate saver for a given base model and discuss the possible extension of this method.
\subsection{Analysis in the context of EEs for performance efficiency}
In EE-based methods, whether to exit after a specific layer or not can be controlled by many factors. Confidence-based exiting is one of the easiest to implement. Here, EEs are designed to fulfil two primary functions: (1) Generate a prediction $\tilde{\boldsymbol{y}}_{\mathit{S}}$ that aligns with the format of the base model's output. (2) Providing information for the evaluation of exiting policy. In classification tasks, this information can be a confidence value $c$ to compare against a pre-configured threshold value $t$. If $c\geq t$, the prediction is deemed acceptable, and the computation process concludes. Otherwise, the computation continues with the remaining layers and finally, the prediction from the base model is accepted $\tilde{\boldsymbol{y}}_{\mathit{B}}$. Alternative exiting policies include but not be limited to stopping when output entropy is low enough\cite{teerapittayanon_branchynet_2017}, consecutive exits yield consistent predictions\cite{zhu_leebert_2021}, or a trained controller decides to exit\cite{dai_epnet_2020}. Trained for the same task, a tiny model can perform identically to the base model. Thus, complete tiny models present no significant obstacles to satisfy both requirements and can be integrated into an early exiting system.

\subsubsection{Eligibility}

The maximal computational savings are acquired when EEs can perform tasks equivalently to larger models. As depicted in \cref{fig:simple-case}(a), conventionally attached exits utilise the backbone to minimize overhead, yet they often struggle to achieve high performance. This is primarily due to the fact that early-stage features in these branches are neither designed nor trained for making predictions at such junctures. In contrast, efficiency-focused tiny models are complete, having all necessary stages for output generation do not encounter this issue. \cref{fig:cn2t-sd} illustrates the comparison between attached exits at different positions in varied scales. However, none of attached exits can outperform independent models with the same level of computation.

Moreover, even some specific co-designs somehow surpass independently developed models in terms of efficiency, the coupled design make them less adaptable to the rapid evolution of modern models. Utilizing widely available efficient model designs can significantly reduce the resources required to build an EE-based system. A straightforward cascading of two models, combined with threshold tuning for early exits, can effectively establish a system with substantial computational savings. These attributes render TinySaver a feasible compression approach for a broad range of applications.

\subsubsection{Efficiency}

We can quantitatively evaluate the efficiency of TinySaver in an EE-based system with two exits. The dataset can be divided into two categories: (I) samples meeting the exit criteria and processed by the saver $\mathit{S}$, and (II) those processed by the base model $\mathit{B}$. Cat. I, filtered through exit criteria, is confidently predicted by $\mathit{S}$, which generally indicates reliable performance. High confidence in predictions typically correlates with reliability, as evidenced in \cref{fig:match-performance}. Here, we examine varying confidence threshold values to chart the performance of both $\mathit{S}$ and base models for Cat. I samples. Initially, when $t$ is large the Cat. I samples are limited to very easy ones, both models exhibit very high performance on them. As the proportion of early exited samples increases, $\mathit{S}$ maintains a relatively high accuracy, while $\mathit{B}$'s performance declines more rapidly. Although large models generally outperform smaller models on average, this does not imply superiority in every individual sample. Given that these samples are selected based on $\mathit{S}$'s confidence, it's not surprising that $\mathit{S}$ may perform better on Cat. I samples. $\mathit{S}$'s performance eventually diminishes as it handles more samples, but there remains a range where results for Cat. I samples are enhanced. Since Cat. II samples continue to be processed by $\mathit{B}$, the system's performance on this segment remains equivalent to that of the large model. Thus, improvements in Cat. I directly contribute to the overall enhancement of the system.
\begin{floatingfigure}[r]{0.5\linewidth}
  \centering
  \includegraphics[width=0.5\linewidth]{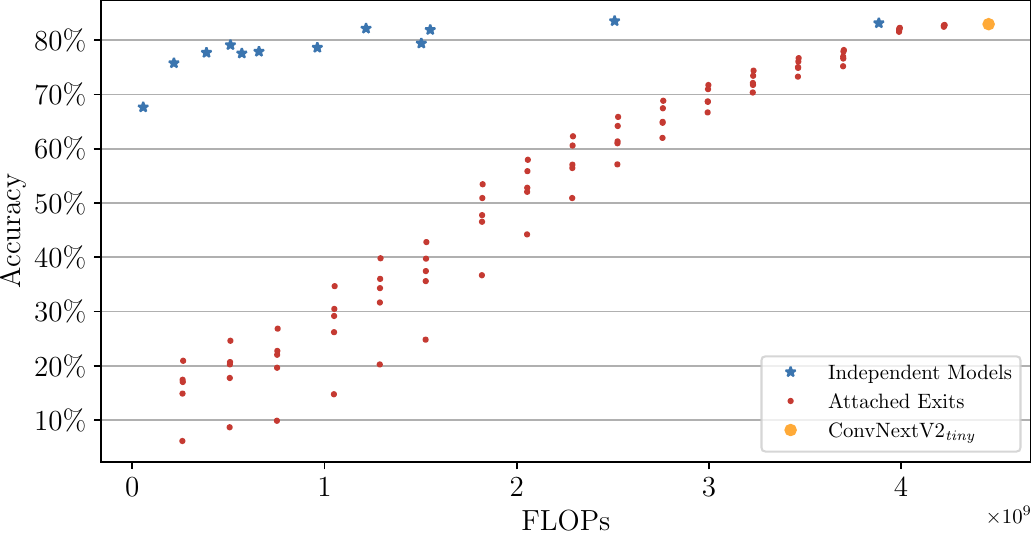}
  \caption{Exits attached on ConvNeXtv2$_\mathit{tiny}$ vs independent models in the similar range of complexity. Exits are attached at various backbone positions, and have different sizes. However, efficient-focused models can easily outperform them. }
  \label{fig:cn2t-sd}
\end{floatingfigure}
The cost factor must also be considered. Cat. II samples incur additional overhead, as they require initial processing by $\mathit{S}$. In contrast, Cat. I samples benefit significantly in cost reduction, bypassing $\mathit{B}$ entirely. For effective compression, the overall complexity should remain below the original complexity of $\mathit{B}$, as indicated before the crossing point in \cref{fig:match-performance}. Achieving this requires a highly efficient saver model. Fortunately, there are many work focusing on creating efficient models. They offer a variety of options for selecting the most suitable saver model in relation to a specific base model.

\subsection{Analysis in the context of MoE for routing efficiency}
As illustrated in \cref{fig:simple-case}(c), this work can be viewed as a variant of the top-1 sparse gated Mixture of Experts (MoE) architecture, featuring two heterogeneous experts, $\mathit{S}$ and $\mathit{B}$ with a top-1 router. The router assess the raw input $\boldsymbol{x}$ to determine which expert to engage. Actually, if the output of $\mathit{S}$ yields a confidence related to per sample performance, it could effectively function as the router in a MoE framework. Our subsequent analysis shows that this model is a competitive alternative to a trained router in complexity reduction scenario.

\subsubsection{Eligibility}
To reduce computation, the router should activate the smaller expert, $\mathit{S}$, when sufficient. Therefore the router must predict the reliability of the result of $\mathit{S}$ for a given input $\boldsymbol{x}$. In classification, that is $\mathit{P}(\mathop{argmax}_{i}(\tilde{\boldsymbol{y}}^{(m,i)}_{\mathit{S}})=\boldsymbol{y}^{(m)}
  \mid \boldsymbol{x}^{(m)})$. However, this formula has the same meaning with $\mathop{max}_{i}(\tilde{\boldsymbol{y}}^{(m,i)}_{\mathit{S}})$, which is also the probability of the prediction matches the ground truth. That implies a classification model $\mathit{S}$ inherently performs the routing function, as its confidence level can serve as a prediction of accuracy. This concept extends to tasks where confidence can be derived from outputs, rendering routers redundant as the two models can inherently form an MoE-like system for computational efficiency. However, an pre or post evaluator is needed for outputs lacking inherent confidence, such as in regression tasks.

\subsubsection{Efficiency}
\label{sec:moe-efficiency}

In TinySaver, computational complexity is reduced to $\mathit{S}$ only, for successfully early exited samples. However, samples failing to meet early exit criteria incur the combined complexities of both $\mathit{S}$ and $\mathit{B}$, as they progress through $\mathit{S}$ before $\mathit{B}$. Thus, the expectation of TinySaver's computational complexity, $E[C_{\mathit{Ts}}]$ can be quantified by \cref{eq:c_new}, where $C_{\mathit{S}}$ and $C_{\mathit{B}}$ represent computational costs, e.g. FLOPs or MACs. $r_{\mathit{S}}$ is the ratio of samples exited via $\mathit{S}$\\
{\setlength{\abovedisplayskip}{2pt}
\setlength{\belowdisplayskip}{0pt}
\centerline{
\resizebox{0.8\linewidth}{!}{
  \begin{minipage}[b]{\linewidth}
    \centering
    \begin{align}
       & E[C_{\mathit{Ts}}] = r_{\mathit{S}} C_{\mathit{S}}+(1-r_{\mathit{S}})(C_{\mathit{B}}+C_{\mathit{S}}) \label{eq:c_new}
    \end{align}
  \end{minipage}
}\\
}}

In a conventional MoE setup with a prior router, there is potential to bypass $\mathit{S}$ for difficult inputs, theoretically saving computations. However, this router must handle raw inputs, necessitating a basic encoder. It makes the cost of the router, $C_{\mathit{Rt}}$, include the costs of an encoder besides the linear and softmax layer. Then the expectation of systematic complexity can be formulated as:\\
{\setlength{\abovedisplayskip}{2pt}
\setlength{\belowdisplayskip}{0pt}
\centerline{
\resizebox{0.8\linewidth}{!}{
  \begin{minipage}[b]{\linewidth}
    \centering
    \begin{align}
       & E[C_{\mathit{MoE}}] = r_{\mathit{S}} \left(C_{\mathit{Rt}}+C_{\mathit{S}} \right) +(1-r_{\mathit{S}})\left(C_{\mathit{Rt}}+C_{\mathit{B}}\right) \label{eq:c_moe}
    \end{align}
  \end{minipage}
}\\
}}
Compare \cref{eq:c_moe} with the proposed method in \cref{eq:c_new}:\\
{\setlength{\abovedisplayskip}{2pt}
\setlength{\belowdisplayskip}{0pt}
\centerline{
\resizebox{0.8\linewidth}{!}{
  \begin{minipage}[b]{\linewidth}
    \centering
    \begin{align}
      E[C_{\mathit{MoE}}]                                                                                                          & <E[C_{\mathit{Ts}}]                                                                                     \\
      r_{\mathit{S}2} \left(C_{\mathit{Rt}}+C_{\mathit{S}} \right) +(1-r_{\mathit{S}2})\left(C_{\mathit{Rt}}+C_{\mathit{B}}\right) & <r_{\mathit{S}1} C_{\mathit{S}}+(1-r_{\mathit{S}1})(C_{\mathit{B}}+C_{\mathit{S}})                      \\
      C_{\mathit{Rt}}                                                                                                              & < (r_{\mathit{S}2}-r_{\mathit{S}1})  C_\mathit{B}+(1-r_{\mathit{S}2}) C_\mathit{S} \label{eq:c_compare}
    \end{align}
  \end{minipage}
}\\
}}

Where $r_{\mathit{S}1}$ and $r_{\mathit{S}2}$ are the exiting ratio of $\mathit{S}$ for each system when they are achieving the equivalent performance. \cref{eq:c_compare} suggests that, the trained router must exhibit significantly lower complexity for the identical routing performance, i.e. $r_{\mathit{S}1}=r_{\mathit{S}2}$, to obtain advantages. In pure classification tasks, the router's learning objective closely align with the classifier, making it improbable for the trained router to offer superior routing with reduced complexity. However, for complex tasks, such as object detection involving bounding box regression, a trained router may provide more efficiency by concentrating on the specific routing task. This is a known limitation of using simple tiny model as-is. Nevertheless, placing an evaluator on the top of $\mathit{S}$'s final layer might be a solution while preserving most features of TinySaver. This post-evaluator can share the backbone with $\mathit{S}$ and introduce limited overhead to plain TinySaver approach.

In summary, employing $\mathit{S}$ as the router offers clear benefits over trained routers in classification tasks. For other types of tasks, a case-by-case analysis is necessary, guided by the conditions in \cref{eq:c_compare}. Moreover, utilizing a pre-trained model as a router significantly reducing deployment and upgrade costs. Consequently, a simple tiny model emerges as a competitive option for computational savings in the context of MoE.

\subsection{Saver model selection}
\label{sec:saver-selection}

\begin{floatingfigure}[r]{0.5\linewidth}
  \centering
  \includegraphics[width=0.5\linewidth]{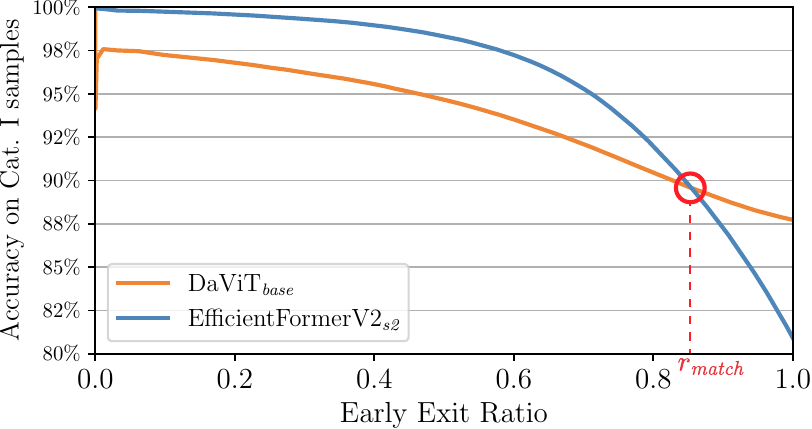}
  \caption{Performance vs early exit ratio on the ImageNet-1k training set. The crossing point (marked in red circle) denotes the estimation of how much base model's job that can be replaced by $\mathit{S}$ model without influencing the systematic performance.}
  \label{fig:match-performance}
\end{floatingfigure}
We propose a method that estimates the compatibility between $\mathit{B}$ and $\mathit{S}$ based on a computation reduction metric $\Delta C_{tr}$ on the training dataset. As illustrated in \cref{fig:match-performance}, the performance of $\mathit{B}$ tends to decline more rapidly as the exit threshold is progressively lowered to contain more samples in Cat. I (Reflecting as the increase of the early exit ratio in \cref{fig:match-performance}). At a certain point, the performance curves of $\mathit{S}$ and $\mathit{B}$ intersect, indicating equivalent performance on the early exited samples. This intersection represents the optimal job ratio $r_{\mathit{match}}$ that $\mathit{S}$ model can undertake without compromising performance. This methodology is encapsulated in \cref{eq:selection-general}, where $m \in \{0, 1, ..., M \}$ denoting the index of each sample in a dataset consisting of $M$ samples. $r_{\mathit{match}}$ is maximized through minimizing $t$, such that the performance of $\mathit{S}$ is no less than $\mathit{B}$.\\
{\setlength{\abovedisplayskip}{2pt}
\setlength{\belowdisplayskip}{0pt}
\centerline{
\resizebox{0.8\linewidth}{!}{
  \begin{minipage}[b]{\linewidth}
    \centering
    \begin{align}
      \begin{split}
        r_{\mathit{match}} = & \mathop{max}_{t}\frac{\# \{m: \mathit{Conf}^{(m)}_{\mathit{S}})\ge t\}}{M} \\
                             & s.t. ~~ \mathit{Perf}_{\mathit{S}}  \geq \mathit{Perf}_{\mathit{B}}
        \label{eq:selection-general}
      \end{split}
    \end{align}
  \end{minipage}
}\\
}}

Specifically, for the classification task, this equation can be formulated in \cref{eq:selection-class} taking into account the predictions made by both $\mathit{S}$ and $\mathit{B}$ for each sample, denoted as $\tilde{\boldsymbol{y}}^{(m,i)}_{\mathit{S}}$ and $\tilde{\boldsymbol{y}}^{(m,i)}_{\mathit{B}}$, respectively. Where $i$ is the index of $N_c$ classes, $i \in \{0, 1, ..., N_c \}$. The confidence level associated with each prediction is derived from the maximum value in the probability vector, which can be expressed as $\mathop{max}_{i}(\tilde{\boldsymbol{y}}^{(m,i)}_{\mathit{S}})$. Here, $r_{\mathit{match}}$ is the maximized ratio of samples that $\mathit{S}$ can take without incurring systematic performance loss.\\
{\setlength{\abovedisplayskip}{2pt}
\setlength{\belowdisplayskip}{0pt}
\centerline{
\resizebox{0.8\linewidth}{!}{
  \begin{minipage}[b]{\linewidth}
    \centering
    \begin{align}
      \begin{split}
        r_{\mathit{match}} = & \mathop{max}_{t}\frac{\# \{m: \mathop{max}_{i}(\tilde{\boldsymbol{y}}^{(m,i)}_{\mathit{S}})\ge t\}}{M} \\ \label{eq:selection-class}
                             & s.t. ~~ \#\{m: \mathop{argmax}_{i}(\tilde{\boldsymbol{y}}^{(m,i)}_{\mathit{S}})=\boldsymbol{y}^{(m)}\} \\ &\geq \#\{m: \mathop{argmax}_{i}(\tilde{\boldsymbol{y}}^{(m,i)}_{\mathit{B}})=\boldsymbol{y}^{(m)}\}
      \end{split}
    \end{align}
  \end{minipage}
}\\
}}
Then the change of complexity, normalized by the complexity of $\mathit{B}$, $\Delta C$ is computed by comparing \cref{eq:c_new} with the standalone complexity $C_{\mathit{B}}$. This $\Delta C$ is also the computation saving ratio for exiting system.
\\
{\setlength{\abovedisplayskip}{2pt}
\setlength{\belowdisplayskip}{0pt}
\centerline{
\resizebox{0.8\linewidth}{!}{
  \begin{minipage}[b]{\linewidth}
    \centering
    \begin{align}
      \Delta C & =\frac{ E[C_{\mathit{Ts}}]  - C_{\mathit{B}}}{C_{\mathit{B}}}\label{eq:delta_c}                                                                                    \\
               & =\frac{1}{C_{\mathit{B}}} \left( r_{\mathit{match}} C_{\mathit{S}}+(1-r_{\mathit{match}}) \label{eq:delta_c2}(C_{\mathit{B}}+C_{\mathit{S}})-C_{\mathit{B}}\right) \\
               & =\frac{C_{\mathit{S}}}{C_{\mathit{B}}}-r_{\mathit{match}}\label{eq:delta_c3}
    \end{align}
  \end{minipage}
}\\
}}

We can apply the aforementioned process on the training set to accurately determine the extent of computational savings, i.e. $\Delta C_{tr}$. Then the metric $\Delta C_{tr}$ serves as an indicator of the efficacy with which saver model collaborates with the base model. $\Delta C_{tr}<0$ signifies that $\mathit{S}$ can compress $\mathit{B}$ without any performance loss on the training set. This calculation method provides a evaluation of the computational change resulting from the integration of the saver and base models within an EE framework.

However, it is important to note the gap between training and testing environments. To ensure reliable compression in practical applications, $\Delta C_{tr}$ should be much lower than 0 to guarantee the compression in practice. Additionally, data augmentation should also be enabled while collecting predictions on the training set to prevent saver from being over-confident.\\

\subsection{Extension with early exit sequence}

Pre-trained tiny models are more efficient than attached exits. However, traditional EE methods possess notable attributes, such as the ability to reuse the backbone, access to early features, be trained specifically, and the potential for multitasking. These flexibilities are not retained with a single tiny model. To address this, we propose the concept of Exit Sequence Network (ESN). This network is designed to collects features provided by the saver model and establish them as a foundation for exits attached to the backbone of the base model. This approach preserves all the benefits of the early exit strategy and further integrates the saver model into the existing network. It enables intermediate exits to merge progressively refined features from the base model with high-level features from the saver model, offering trade-offs at different levels in predictions. Similar to other EE-based models, the heads built on the ESN is not necessarily having the same task as the original model, providing rich flexibility. However, ESN uniquely includes the tiny model, ensuring a lower bound of performance. The detailed design considerations for ESN are discussed in the Appendix \cref{sec:esn_detail}. Additionally, as ESN integrates the concept of TinySaver into traditional EE contexts,  it allows us evaluate the impact of incorporating an independent tiny model within an EE-based system.

\begin{figure}[h]
  \centering
  \includegraphics[width=1\linewidth]{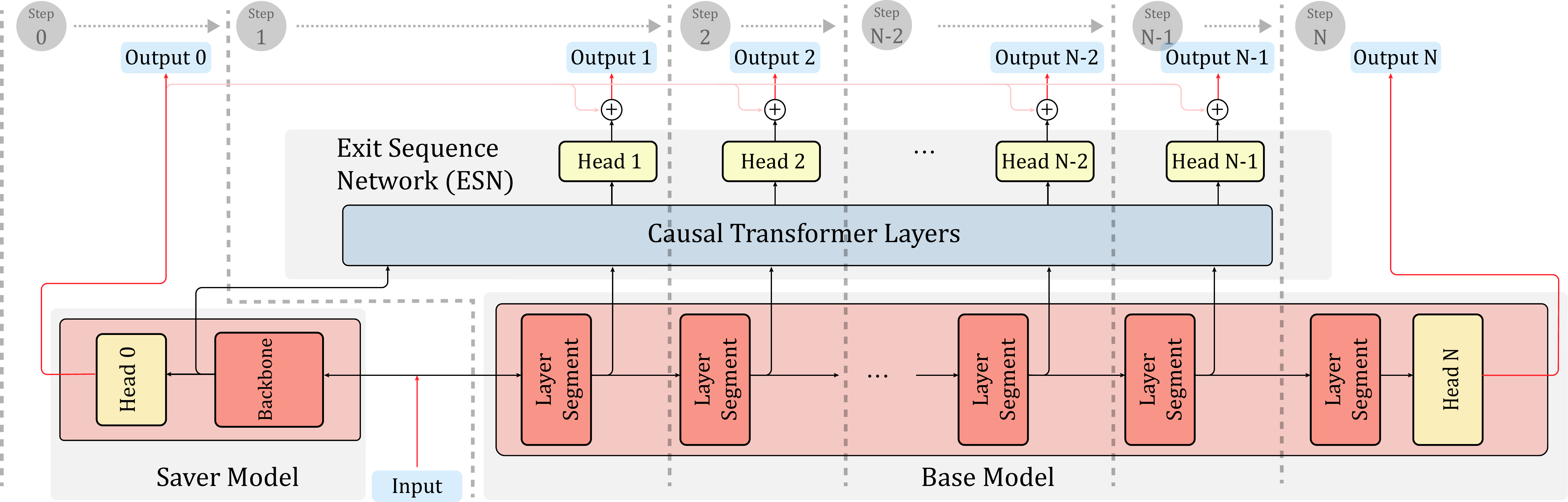}
  \caption{TinySaver with the Exit Sequence Network}
  \label{fig:esn}
\end{figure}

\section{Experiment}

In this section, we present the comprehensive analysis of TinySaver with its compression impact on a diverse set of vision models, mainly focusing on the ImageNet-1k\cite{russakovsky_imagenet_2015} classification task. Additionally, we explore the saver selection methodology and the ESN framework. Our experiments encompass a pool of 44 models, each capable of utilizing another model as its saver. Detailed descriptions of these models, experimental configurations and the complete results are documented in the appendix \cref{sec:exp-details}. Moreover, we extend our validate to the object detection task on the COCO dataset\cite{lin_microsoft_2015}.
\subsection{Compression of general models}
\label{sec:results-main}
We conduct simulations to assess TinySaver's efficiency in compressing general models. Inference is performed on the training set, incorporating data augmentation techniques\cite{cubuk_autoaugment_2019}, and we log all predictions and associated confidences from a variety of pre-trained models. Subsequently, for each model, we identify a smaller model with the lowest $\Delta C_{tr}$ to act as its saver. Following this selection, we test the system on the validation set across varying confidence thresholds $t \in [0,1]$.

The simulation results for eight models, split between four large and four small, are illustrated in \cref{tab:result-main}. They demonstrate TinySaver's effectiveness in significantly reducing overall FLOPs, particularly in larger models. Most models experienced reductions in complexity without sacrificing performance. These models also receive considerable savings while incurring negligible accuracy losses, especially for four large models. The TinySaver even makes some models surpass their original performance with certain threshold settings. Notably, all selected savers are efficiency-focused designs, rather than simply being scaled-down versions of high-performance architectures. This suggests that despite many SOTA works claiming scalability, efficiency-focused models may offer even superior performance at reduced scales.

\begin{table*}[ht]
   \caption{Compression effect while TinySaver applied to models at various architectures and scales. Evaluation is performed on IN-1K classification.}
   \label{tab:result-main}

   \resizebox{\textwidth}{!}{%
      \begin{tabular}{lcc|lcc|cccc|cc}
         \hline
         \multicolumn{3}{l|}{Base Model}                             & \multicolumn{3}{l|}{Saver Model} & \multicolumn{4}{c|}{Required GFLOPs(\%) when Acc. Drop $\leq$} & \multicolumn{2}{c}{Max   Performance}                                                                                                                                                                                                                                                                                                                                                                                                                                                                                                                                                                                 \\
         Name                                                        & GFLOPs.                          & Acc.                                                           & Name                                                        & GFLOPs. & Acc.    & \multicolumn{1}{c}{1\%}                                                             & \multicolumn{1}{c}{0.5\%}                                                           & \multicolumn{1}{c}{0.1\%}                                                          & \multicolumn{1}{c|}{0\%}                                                            & \multicolumn{1}{r}{Avg. GFLOPs}                                                     & \multicolumn{1}{r}{Acc.}                                                              \\ \hline
         \rowcolor[gray]{0.95}
         ConvNeXtV2$_\mathit{huge}$\cite{woo_convnext_2023-1}        & 114.92                           & 86.25\%                                                        & EfficientFormerV2$_\mathit{l}$\cite{li_rethinking_2023}     & 2.51    & 83.53\% & \begin{tabular}[c]{@{}l@{}}13.21\\      \textcolor{mygreen}{(-88.5\%)}\end{tabular} & \begin{tabular}[c]{@{}l@{}}19.41\\      \textcolor{mygreen}{(-83.1\%)}\end{tabular} & \begin{tabular}[c]{@{}l@{}}31.17\\      \textcolor{mygreen}{(-72.9\%)}\end{tabular} & \begin{tabular}[c]{@{}l@{}}67.34\\      \textcolor{mygreen}{(-41.4\%)}\end{tabular} & \begin{tabular}[c]{@{}l@{}}70.70\\      \textcolor{mygreen}{(-38.5\%)}\end{tabular} & \begin{tabular}[c]{@{}l@{}}86.25\%\\      \textcolor{mygreen}{(+0.00\%)}\end{tabular} \\
         MaxViT$_\mathit{large}$\cite{tu_maxvit_2022}                & 42.80                            & 84.83\%                                                        & EfficientFormerV2$_\mathit{l}$\cite{li_rethinking_2023}     & 2.51    & 83.53\% & \begin{tabular}[c]{@{}l@{}}3.46\\      \textcolor{mygreen}{(-91.9\%)}\end{tabular}  & \begin{tabular}[c]{@{}l@{}}5.48\\      \textcolor{mygreen}{(-87.2\%)}\end{tabular}  & \begin{tabular}[c]{@{}l@{}}7.81\\      \textcolor{mygreen}{(-81.8\%)}\end{tabular}  & \begin{tabular}[c]{@{}l@{}}9.04\\      \textcolor{mygreen}{(-78.9\%)}\end{tabular}  & \begin{tabular}[c]{@{}l@{}}11.24\\      \textcolor{mygreen}{(-73.7\%)}\end{tabular} & \begin{tabular}[c]{@{}l@{}}84.92\%\\      \textcolor{mygreen}{(+0.09\%)}\end{tabular} \\
         \rowcolor[gray]{0.95}
         DaViT$_\mathit{base}$\cite{ding_davit_2022}                 & 15.22                            & 84.49\%                                                        & EfficientFormerV2$_\mathit{s2}$\cite{li_rethinking_2023}    & 1.22    & 82.16\% & \begin{tabular}[c]{@{}l@{}}2.40\\      \textcolor{mygreen}{(-84.3\%)}\end{tabular}  & \begin{tabular}[c]{@{}l@{}}3.16\\      \textcolor{mygreen}{(-79.2\%)}\end{tabular}  & \begin{tabular}[c]{@{}l@{}}4.40\\      \textcolor{mygreen}{(-71.1\%)}\end{tabular}  & \begin{tabular}[c]{@{}l@{}}5.19\\      \textcolor{mygreen}{(-65.9\%)}\end{tabular}  & \begin{tabular}[c]{@{}l@{}}6.50\\      \textcolor{mygreen}{(-57.3\%)}\end{tabular}  & \begin{tabular}[c]{@{}l@{}}84.53\%\\      \textcolor{mygreen}{(+0.04\%)}\end{tabular} \\
         Swin$_\mathit{base}$\cite{liu_swin_2021-1}                  & 15.13                            & 85.15\%                                                        & EfficientFormerV2$_\mathit{s2}$\cite{li_rethinking_2023}    & 1.22    & 82.16\% & \begin{tabular}[c]{@{}l@{}}2.81\\      \textcolor{mygreen}{(-81.4\%)}\end{tabular}  & \begin{tabular}[c]{@{}l@{}}3.70\\      \textcolor{mygreen}{(-75.6\%)}\end{tabular}  & \begin{tabular}[c]{@{}l@{}}4.79\\      \textcolor{mygreen}{(-68.3\%)}\end{tabular}  & \begin{tabular}[c]{@{}l@{}}9.06\\      \textcolor{mygreen}{(-40.1\%)}\end{tabular}  & \begin{tabular}[c]{@{}l@{}}9.40\\      \textcolor{mygreen}{(-37.9\%)}\end{tabular}  & \begin{tabular}[c]{@{}l@{}}85.15\%\\      \textcolor{mygreen}{(+0.00\%)}\end{tabular} \\
         \rowcolor[gray]{0.95}
         EfficientNet$_\mathit{b2}$\cite{tan_efficientnet_2019-1}    & 0.66                             & 77.90\%                                                        & MobileNetV3$_\mathit{large100}$\cite{howard_searching_2019} & 0.22    & 75.78\% & \begin{tabular}[c]{@{}l@{}}0.27\\      \textcolor{mygreen}{(-59.4\%)}\end{tabular}  & \begin{tabular}[c]{@{}l@{}}0.30\\      \textcolor{mygreen}{(-53.9\%)}\end{tabular}  & \begin{tabular}[c]{@{}l@{}}0.34\\      \textcolor{mygreen}{(-47.8\%)}\end{tabular}  & \begin{tabular}[c]{@{}l@{}}0.36\\      \textcolor{mygreen}{(-45.5\%)}\end{tabular}  & \begin{tabular}[c]{@{}l@{}}0.46\\      \textcolor{mygreen}{(-30.9\%)}\end{tabular}  & \begin{tabular}[c]{@{}l@{}}78.29\%\\      \textcolor{mygreen}{(+0.39\%)}\end{tabular} \\
         EfficientViT$_\mathit{m5}$\cite{liu_efficientvit_2023}      & 0.52                             & 77.08\%                                                        & MobileNetV3$_\mathit{large100}$\cite{howard_searching_2019} & 0.22    & 75.78\% & \begin{tabular}[c]{@{}l@{}}0.23\\      \textcolor{mygreen}{(-55.7\%)}\end{tabular}  & \begin{tabular}[c]{@{}l@{}}0.25\\      \textcolor{mygreen}{(-51.2\%)}\end{tabular}  & \begin{tabular}[c]{@{}l@{}}0.28\\      \textcolor{mygreen}{(-44.9\%)}\end{tabular}  & \begin{tabular}[c]{@{}l@{}}0.29\\      \textcolor{mygreen}{(-43.0\%)}\end{tabular}  & \begin{tabular}[c]{@{}l@{}}0.38\\      \textcolor{mygreen}{(-27.1\%)}\end{tabular}  & \begin{tabular}[c]{@{}l@{}}77.55\%\\      \textcolor{mygreen}{(+0.47\%)}\end{tabular} \\
         \rowcolor[gray]{0.95}
         EfficientFormerV2$_\mathit{s0}$\cite{li_rethinking_2023}    & 0.38                             & 76.25\%                                                        & EfficientViT$_\mathit{b0}$\cite{cai_efficientvit_2023}      & 0.10    & 71.35\% & \begin{tabular}[c]{@{}l@{}}0.18\\      \textcolor{mygreen}{(-51.0\%)}\end{tabular}  & \begin{tabular}[c]{@{}l@{}}0.20\\      \textcolor{mygreen}{(-46.0\%)}\end{tabular}  & \begin{tabular}[c]{@{}l@{}}0.22\\      \textcolor{mygreen}{(-41.3\%)}\end{tabular}  & \begin{tabular}[c]{@{}l@{}}0.23\\      \textcolor{mygreen}{(-38.8\%)}\end{tabular}  & \begin{tabular}[c]{@{}l@{}}0.26\\      \textcolor{mygreen}{(-30.6\%)}\end{tabular}  & \begin{tabular}[c]{@{}l@{}}76.38\%\\      \textcolor{mygreen}{(+0.13\%)}\end{tabular} \\
         MobileNetV3$_\mathit{large100}$\cite{howard_searching_2019} & 0.22                             & 75.78\%                                                        & MobileNetV3$_\mathit{small100}$\cite{howard_searching_2019} & 0.06    & 67.65\% & \begin{tabular}[c]{@{}l@{}}0.14\\      \textcolor{mygreen}{(-33.6\%)}\end{tabular}  & \begin{tabular}[c]{@{}l@{}}0.16\\      \textcolor{mygreen}{(-26.5\%)}\end{tabular}  & \begin{tabular}[c]{@{}l@{}}0.19\\      \textcolor{mygreen}{(-12.2\%)}\end{tabular}  & \begin{tabular}[c]{@{}l@{}}0.22\\      \textcolor{myred}{(+2.8\%)}\end{tabular}     & \begin{tabular}[c]{@{}l@{}}0.22\\      \textcolor{myred}{(+3.4\%)}\end{tabular}     & \begin{tabular}[c]{@{}l@{}}75.78\%\\      \textcolor{mygreen}{(+0.00\%)}\end{tabular} \\
         \hline
      \end{tabular}%
   }

\end{table*}

\subsection{$\Delta C_{tr}$ and validation performance}

To validate our saver selection approach, we conducted simulations by applying various savers to the same base model, with outcomes presented in \cref{tab:result-large}. Due to the difference between the training and validation set, the saver with the smallest $\Delta C_{tr}$ does not invariably equate to the best validation performance. Nonetheless, a high correlation exists between lower $\Delta C_{tr}$ and greater computational savings, indicating that this metric can effectively and quickly identify promising saver options in practice.

\begin{table}[h]
   \begin{minipage}[c]{0.48\linewidth}

      \caption{Compression when different savers are selected. Lower $\Delta C_{tr}$ correlates with better results. Best cases are marked in bold.}
      \label{tab:result-large}
      \resizebox{\linewidth}{!}{
         \begin{tabular}{l|l|ccc|l}
            \hline
                                                                                       &                                                                                   & \multicolumn{3}{c|}{Required GFLOPs(\%) when Acc. Drop $\leq$}                                                  & \multicolumn{1}{r}{}                                                                                                                                                                                                                                                                 \\
            \multirow{-2}{*}{Base Model}                                               & \multirow{-2}{*}{Saver Model}                                                     & 1\%                                                                                                             & 0.1\%                                                                                                          & 0\%                                                                                                             & \multicolumn{1}{r}{\multirow{-2}{*}{$\Delta C$}} \\ \hline
                                                                                       & \cellcolor[gray]{0.95}EfficientFormerV2$_\mathit{l}$\cite{li_rethinking_2023}     & \cellcolor[gray]{0.95}\textbf{\begin{tabular}[c]{@{}c@{}}3.47      \textcolor{mygreen}{(-91.9\%)}\end{tabular}} & \cellcolor[gray]{0.95}\textbf{\begin{tabular}[c]{@{}c@{}}7.81      \textcolor{mygreen}{(-81.8\%)}\end{tabular}} & \cellcolor[gray]{0.95}\textbf{\begin{tabular}[c]{@{}c@{}}9.31      \textcolor{mygreen}{(-78.2\%)}\end{tabular}} & \cellcolor[gray]{0.95}\textbf{-0.81}             \\
                                                                                       & EfficientViT$_\mathit{b3}$\cite{cai_efficientvit_2023}                            & \begin{tabular}[c]{@{}c@{}}6.11      \textcolor{mygreen}{(-85.7\%)}\end{tabular}                                & \begin{tabular}[c]{@{}c@{}}10.92      \textcolor{mygreen}{(-74.5\%)}\end{tabular}                               & \begin{tabular}[c]{@{}c@{}}13.50      \textcolor{mygreen}{(-68.5\%)}\end{tabular}                               & -0.79                                            \\
                                                                                       & \cellcolor[gray]{0.95}DaViT$_\mathit{tiny}$\cite{ding_davit_2022}                 & \cellcolor[gray]{0.95}\begin{tabular}[c]{@{}c@{}}7.71      \textcolor{mygreen}{(-82.0\%)}\end{tabular}          & \cellcolor[gray]{0.95}\begin{tabular}[c]{@{}c@{}}13.78      \textcolor{mygreen}{(-67.8\%)}\end{tabular}         & \cellcolor[gray]{0.95}\begin{tabular}[c]{@{}c@{}}16.41      \textcolor{mygreen}{(-61.7\%)}\end{tabular}         & \cellcolor[gray]{0.95}-0.77                      \\
                                                                                       & EfficientFormerV2$_\mathit{s2}$\cite{li_rethinking_2023}                          & \begin{tabular}[c]{@{}c@{}}4.89      \textcolor{mygreen}{(-88.6\%)}\end{tabular}                                & \begin{tabular}[c]{@{}c@{}}11.22      \textcolor{mygreen}{(-73.8\%)}\end{tabular}                               & \begin{tabular}[c]{@{}c@{}}12.80      \textcolor{mygreen}{(-70.1\%)}\end{tabular}                               & -0.77                                            \\
            \multirow{-5}{*}{MaxViT$_\mathit{large}$\cite{tu_maxvit_2022}}             & \cellcolor[gray]{0.95}ConvNeXtV2$_\mathit{tiny}$\cite{woo_convnext_2023-1}        & \cellcolor[gray]{0.95}\begin{tabular}[c]{@{}c@{}}7.18      \textcolor{mygreen}{(-83.2\%)}\end{tabular}          & \cellcolor[gray]{0.95}\begin{tabular}[c]{@{}c@{}}12.18      \textcolor{mygreen}{(-71.5\%)}\end{tabular}         & \cellcolor[gray]{0.95}\begin{tabular}[c]{@{}c@{}}15.06      \textcolor{mygreen}{(-64.8\%)}\end{tabular}         & \cellcolor[gray]{0.95}-0.76                      \\ \hline
                                                                                       & EfficientFormerV2$_\mathit{s2}$\cite{li_rethinking_2023}                          & \textbf{\begin{tabular}[c]{@{}c@{}}2.40      \textcolor{mygreen}{(-84.3\%)}\end{tabular}}                       & \begin{tabular}[c]{@{}c@{}}4.40      \textcolor{mygreen}{(-71.1\%)}\end{tabular}                                & \begin{tabular}[c]{@{}c@{}}5.19      \textcolor{mygreen}{(-65.9\%)}\end{tabular}                                & \textbf{-0.78}                                   \\
                                                                                       & \cellcolor[gray]{0.95}EfficientFormerV2$_\mathit{l}$\cite{li_rethinking_2023}     & \cellcolor[gray]{0.95}-                                                                                         & \cellcolor[gray]{0.95}\textbf{\begin{tabular}[c]{@{}c@{}}3.96      \textcolor{mygreen}{(-74.0\%)}\end{tabular}} & \cellcolor[gray]{0.95}\textbf{\begin{tabular}[c]{@{}c@{}}4.19      \textcolor{mygreen}{(-72.4\%)}\end{tabular}} & \cellcolor[gray]{0.95}-0.76                      \\
                                                                                       & EfficientNet$_\mathit{b4}$\cite{tan_efficientnet_2019-1}                          & \begin{tabular}[c]{@{}c@{}}4.86      \textcolor{mygreen}{(-68.1\%)}\end{tabular}                                & \begin{tabular}[c]{@{}c@{}}8.71      \textcolor{mygreen}{(-42.8\%)}\end{tabular}                                & \begin{tabular}[c]{@{}c@{}}15.18      \textcolor{mygreen}{(-0.3\%)}\end{tabular}                                & -0.73                                            \\
                                                                                       & \cellcolor[gray]{0.95}EfficientNet$_\mathit{b3}$\cite{tan_efficientnet_2019-1}    & \cellcolor[gray]{0.95}\begin{tabular}[c]{@{}c@{}}5.06      \textcolor{mygreen}{(-66.7\%)}\end{tabular}          & \cellcolor[gray]{0.95}\begin{tabular}[c]{@{}c@{}}9.35      \textcolor{mygreen}{(-38.6\%)}\end{tabular}          & \cellcolor[gray]{0.95}\begin{tabular}[c]{@{}c@{}}15.22      \textcolor{mygreen}{(-0.0\%)}\end{tabular}          & \cellcolor[gray]{0.95}-0.71                      \\
            \multirow{-5}{*}{DaViT$_\mathit{base}$\cite{ding_davit_2022}}              & ConvNeXtV2$_\mathit{nano}$\cite{woo_convnext_2023-1}                              & \begin{tabular}[c]{@{}c@{}}4.00      \textcolor{mygreen}{(-73.7\%)}\end{tabular}                                & \begin{tabular}[c]{@{}c@{}}6.46      \textcolor{mygreen}{(-57.5\%)}\end{tabular}                                & \begin{tabular}[c]{@{}c@{}}7.98      \textcolor{mygreen}{(-47.6\%)}\end{tabular}                                & -0.7                                             \\ \hline
                                                                                       & \cellcolor[gray]{0.95}MobileNetV3$_\mathit{large100}$\cite{howard_searching_2019} & \cellcolor[gray]{0.95}\textbf{\begin{tabular}[c]{@{}c@{}}0.27      \textcolor{mygreen}{(-59.2\%)}\end{tabular}} & \cellcolor[gray]{0.95}\textbf{\begin{tabular}[c]{@{}c@{}}0.35      \textcolor{mygreen}{(-47.5\%)}\end{tabular}} & \cellcolor[gray]{0.95}\textbf{\begin{tabular}[c]{@{}c@{}}0.36      \textcolor{mygreen}{(-45.5\%)}\end{tabular}} & \cellcolor[gray]{0.95}\textbf{-0.58}             \\
                                                                                       & EfficientViT$_\mathit{b0}$\cite{cai_efficientvit_2023}                            & \begin{tabular}[c]{@{}c@{}}0.28      \textcolor{mygreen}{(-57.0\%)}\end{tabular}                                & \begin{tabular}[c]{@{}c@{}}0.38      \textcolor{mygreen}{(-42.6\%)}\end{tabular}                                & \begin{tabular}[c]{@{}c@{}}0.39      \textcolor{mygreen}{(-41.1\%)}\end{tabular}                                & -0.47                                            \\
                                                                                       & \cellcolor[gray]{0.95}MobileNetV3$_\mathit{small100}$\cite{howard_searching_2019} & \cellcolor[gray]{0.95}\begin{tabular}[c]{@{}c@{}}0.34      \textcolor{mygreen}{(-48.5\%)}\end{tabular}          & \cellcolor[gray]{0.95}\begin{tabular}[c]{@{}c@{}}0.48      \textcolor{mygreen}{(-27.2\%)}\end{tabular}          & \cellcolor[gray]{0.95}\begin{tabular}[c]{@{}c@{}}0.53      \textcolor{mygreen}{(-20.0\%)}\end{tabular}          & \cellcolor[gray]{0.95}-0.42                      \\
                                                                                       & EfficientViT$_\mathit{m4}$\cite{liu_efficientvit_2023}                            & \begin{tabular}[c]{@{}c@{}}0.41      \textcolor{mygreen}{(-37.5\%)}\end{tabular}                                & \begin{tabular}[c]{@{}c@{}}0.49      \textcolor{mygreen}{(-26.0\%)}\end{tabular}                                & \begin{tabular}[c]{@{}c@{}}0.51      \textcolor{mygreen}{(-23.0\%)}\end{tabular}                                & -0.4                                             \\
            \multirow{-5}{*}{EfficientNet$_\mathit{b2}$\cite{tan_efficientnet_2019-1}} & \cellcolor[gray]{0.95}EfficientViT$_\mathit{m3}$\cite{liu_efficientvit_2023}      & \cellcolor[gray]{0.95}\begin{tabular}[c]{@{}c@{}}0.41      \textcolor{mygreen}{(-38.2\%)}\end{tabular}          & \cellcolor[gray]{0.95}\begin{tabular}[c]{@{}c@{}}0.50      \textcolor{mygreen}{(-23.9\%)}\end{tabular}          & \cellcolor[gray]{0.95}\begin{tabular}[c]{@{}c@{}}0.54      \textcolor{mygreen}{(-17.5\%)}\end{tabular}          & \cellcolor[gray]{0.95}-0.39                      \\ \hline
         \end{tabular}%
      }
   \end{minipage}\hfill
   \begin{minipage}[c]{0.48\linewidth}
      \caption{TinySaver with ESN enabled. Each base model is employing the same saver as in \cref{tab:result-main}, figures have improvement are marked in bold.}
      \label{tab:esn-enabled}
      \resizebox{\linewidth}{!}{%
         \begin{tabular}{l|l|cccc}
            \hline
            Base Model                                                   & Saver Model                                                  & \multicolumn{4}{c}{Required GFLOPs(\%) when Acc. Drop $\leq$}                                                                                                                                                                                                                                                                                                         \\
            Name                                                         & Name                                                         & 1\%                                                                                     & 0.5\%                                                                                   & 0.1\%                                                                                  & 0\%                                                                                     \\ \hline
            \rowcolor[gray]{0.95}
            ConvNeXtV2$_\mathit{huge}$\cite{woo_convnext_2023-1}         & EfficientFormerV2$_\mathit{l}$\cite{li_rethinking_2023}      & \textbf{\begin{tabular}[c]{@{}c@{}}13.07\\ \textcolor{mygreen}{(-88.6\%)}\end{tabular}} & \begin{tabular}[c]{@{}c@{}}19.67\\ \textcolor{mygreen}{(-82.9\%)}\end{tabular}          & -                                                                                       & -                                                                                       \\
            MaxViT$_\mathit{large}$\cite{tu_maxvit_2022}                 & EfficientFormerV2$_\mathit{l}$\cite{li_rethinking_2023}      & \begin{tabular}[c]{@{}c@{}}3.46\\ \textcolor{mygreen}{(-91.9\%)}\end{tabular}           & \begin{tabular}[c]{@{}c@{}}5.48\\ \textcolor{mygreen}{(-87.2\%)}\end{tabular}           & \begin{tabular}[c]{@{}c@{}}7.79\\ \textcolor{mygreen}{(-81.8\%)}\end{tabular}           & \begin{tabular}[c]{@{}c@{}}9.03\\ \textcolor{mygreen}{(-78.9\%)}\end{tabular}           \\
            \rowcolor[gray]{0.95}
            DaViT$_\mathit{base}$\cite{ding_davit_2022}                  & EfficientFormerV2$_\mathit{s2}$\cite{li_rethinking_2023}     & \begin{tabular}[c]{@{}c@{}}2.40\\ \textcolor{mygreen}{(-84.2\%)}\end{tabular}           & \begin{tabular}[c]{@{}c@{}}3.16\\ \textcolor{mygreen}{(-79.2\%)}\end{tabular}           & \textbf{\begin{tabular}[c]{@{}c@{}}4.38\\ \textcolor{mygreen}{(-71.2\%)}\end{tabular}}  & \textbf{\begin{tabular}[c]{@{}c@{}}5.12\\ \textcolor{mygreen}{(-66.4\%)}\end{tabular}}  \\
            Swin$_\mathit{base}$\cite{liu_swin_2021-1}                   & EfficientFormerV2$_\mathit{s2}$\cite{li_rethinking_2023}     & \textbf{\begin{tabular}[c]{@{}c@{}}2.80\\ \textcolor{mygreen}{(-81.5\%)}\end{tabular}}  & \begin{tabular}[c]{@{}c@{}}3.69\\ \textcolor{mygreen}{(-75.6\%)}\end{tabular}           & \textbf{\begin{tabular}[c]{@{}c@{}}4.72\\ \textcolor{mygreen}{(-68.8\%)}\end{tabular}}  & -                                                                                       \\
            \rowcolor[gray]{0.95}
            EfficientNet$_\mathit{b2}$\cite{tan_efficientnet_2019-1}     & MobileNetV3$_\mathit{large_100}$\cite{howard_searching_2019} & \begin{tabular}[c]{@{}c@{}}0.27\\ \textcolor{mygreen}{(-59.3\%)}\end{tabular}           & \textbf{\begin{tabular}[c]{@{}c@{}}0.30\\ \textcolor{mygreen}{(-54.0\%)}\end{tabular}}  & \begin{tabular}[c]{@{}c@{}}0.34\\ \textcolor{mygreen}{(-47.7\%)}\end{tabular}           & \begin{tabular}[c]{@{}c@{}}0.36\\ \textcolor{mygreen}{(-45.5\%)}\end{tabular}           \\
            EfficientViT$_\mathit{m5}$\cite{liu_efficientvit_2023}       & MobileNetV3$_\mathit{large_100}$\cite{howard_searching_2019} & \textbf{\begin{tabular}[c]{@{}c@{}}0.23\\ \textcolor{mygreen}{(-55.8\%)}\end{tabular} } & \textbf{ \begin{tabular}[c]{@{}c@{}}0.25\\ \textcolor{mygreen}{(-51.3\%)}\end{tabular}} & \textbf{\begin{tabular}[c]{@{}c@{}}0.28\\ \textcolor{mygreen}{(-45.3\%)}\end{tabular}}  & \textbf{\begin{tabular}[c]{@{}c@{}}0.29\\ \textcolor{mygreen}{(-43.2\%)}\end{tabular}}  \\
            \rowcolor[gray]{0.95}
            EfficientFormerV2$_\mathit{s0}$\cite{li_rethinking_2023}     & EfficientViT$_\mathit{b0}$\cite{cai_efficientvit_2023}       & \begin{tabular}[c]{@{}c@{}}0.19\\ \textcolor{mygreen}{(-51.0\%)}\end{tabular}           & \begin{tabular}[c]{@{}c@{}}0.20\\ \textcolor{mygreen}{(-45.8\%)}\end{tabular}           & \begin{tabular}[c]{@{}c@{}}0.22\\ \textcolor{mygreen}{(-41.1\%)}\end{tabular}           & \begin{tabular}[c]{@{}c@{}}0.23\\ \textcolor{mygreen}{(-38.4\%)}\end{tabular}           \\
            MobileNetV3$_\mathit{large_100}$\cite{howard_searching_2019} & MobileNetV3$_\mathit{small_100}$\cite{howard_searching_2019} & \begin{tabular}[c]{@{}c@{}}0.14\\ \textcolor{mygreen}{(-33.6\%)}\end{tabular}           & \begin{tabular}[c]{@{}c@{}}0.16\\ \textcolor{mygreen}{(-26.5\%)}\end{tabular}           & \textbf{\begin{tabular}[c]{@{}c@{}}0.19\\ \textcolor{mygreen}{(-12.7\%)}\end{tabular} } & \textbf{\begin{tabular}[c]{@{}c@{}}0.21\\ \textcolor{mygreen}{(-2.9\%)} \end{tabular} }
            \\
            \hline
         \end{tabular}%
      }

   \end{minipage}
\end{table}

\subsection{Testing with ESN}
In \cref{tab:esn-enabled}, we detail TinySaver's performance when enhanced with ESN, testing it across the same eight models and savers as in \cref{tab:result-main}. We conducted a basic search for optimal hyper-parameters, including the number of attention layers and dimensions of hidden neurons. The listed results reflect the best configurations for different performance goals. Additionally, we employed DyCE\cite{wang_dyce_2024} to tune exit thresholds, with full details in \cref{sec:esn_training_detail}. Our findings reveal no substantial improvements using ESN compared to the plain TinySaver, suggesting that a tiny model is already highly efficient and effective, in classification, making traditional EE architectures unable to provide further significant benefits.

\subsection{Comparison with related work}

\begin{figure}[h]
   \centering
   \subcaptionbox{With EE-based models}{\includegraphics[width=0.5\textwidth]{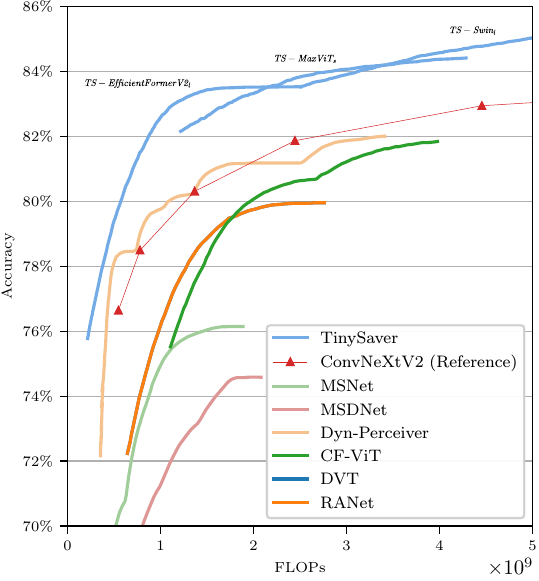}}%
   \hfill
   \subcaptionbox{With Static and MoE-based models}{\includegraphics[width=0.5\textwidth]{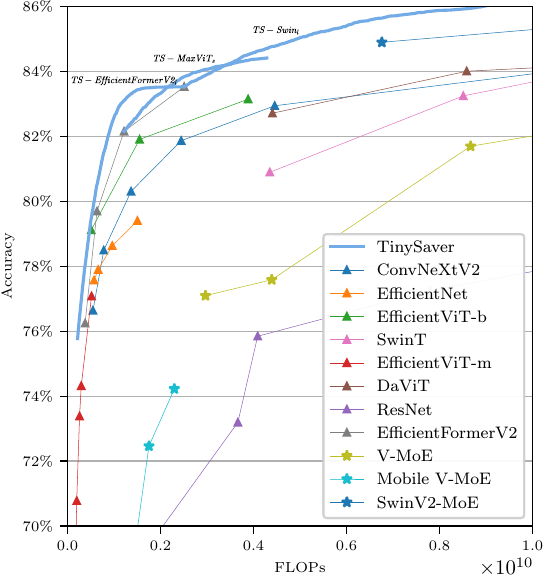}}
   \caption{Comparison of the top-1 accuracy on ImageNet-1k between TinySaver enabled Models and related work}
   \label{fig:comparison}

\end{figure}

In \cref{fig:comparison}, we compare TinySaver against various existing models from EE-based\cite{chen_cf-vit_2022,wang_not_2021,han_dynamic_2023,hang_msnet_2023,hong_panda_2021}, MoE-based\cite{daxberger_mobile_2023,riquelme_scaling_2021,hwang_tutel_2023}, and static categories\cite{ding_davit_2022,woo_convnext_2023-1,tu_maxvit_2022,dosovitskiy_image_2021-1,he_deep_2016,tan_efficientnet_2019-1,liu_efficientvit_2023,cai_efficientvit_2023,li_rethinking_2023}. As presented in \cref{fig:comparison}(a), TinySaver outperforms EE-based methods significantly due to its model-agnostic feature, which allows it to seamlessly leverage improvement across models, datasets, and training recipes, while the progress of other EE-based work relies on refining design gradually. Compared to static models in \cref{fig:comparison}(b), TinySaver surpasses scaling down model architectures directly. It enhances the original model with efficiencies from other models, outperforming smaller versions of the original. The TinySaver enabled model also usually has higher accuracy than all static models in a similar complexity range. Moreover, as a dynamic method which has a continuous performance curve, it is configurable in fine-grain, during runtime to adapt to specific needs. It is also orthogonal to most existing compression methods like quantization. Combining these approaches could yield further efficiency gains. We also include the comparison with MoE-based models in \cref{fig:comparison}(b). Although MoE-based architectures are efficient in scaling models up, they are not as efficient as TinySaver in reducing complexity. All these observations demonstrate the effectiveness of integrating tiny models as computational savers.

\begin{wraptable}{R}{0.5\linewidth}
   \caption{Results of objection detection on COCO 2017val\cite{lin_microsoft_2015}.}
   \label{tab:result-coco}
   \centering

   \begin{adjustbox}{max width=\linewidth}
      \begin{tabular}{llc|cc}
         \hline
         Method                                                              & Model                                                                                                                     & \begin{tabular}[c]{@{}c@{}}Pretrained\\      Dataset\end{tabular} & Box AP & GFLOPs         \\
         \hline
         -                                                                   & YOLOv8n\cite{yolov8_ultralytics}                                                                                          & IN1k                                                              & 37.3   & 9              \\
         -                                                                   & YOLOv8s\cite{yolov8_ultralytics}                                                                                          & IN1k                                                              & 44.9   & 29             \\
         Efficient-DETR\cite{yao_efficient_2021}                             & R50\cite{he_deep_2016}                                                                                                    & IN1k                                                              & 45.1   & 210            \\
         \rowcolor[gray]{0.95}
         TinySaver($t=0.25$)                                                 & \begin{tabular}[l]{@{}l@{}}YOLOv8x\cite{yolov8_ultralytics}\\      +YOLOv8n\cite{yolov8_ultralytics}\end{tabular}         & IN1k                                                              & 45.3   & 71             \\
         -                                                                   & YOLOv8m\cite{yolov8_ultralytics}                                                                                          & IN1k                                                              & 50.2   & 79             \\
         \rowcolor[gray]{0.95}
         TinySaver($t=0.25$)                                                 & \begin{tabular}[l]{@{}l@{}}YOLOv8l\cite{yolov8_ultralytics}\\      +YOLOv8s\cite{yolov8_ultralytics}\end{tabular}         & IN1k                                                              & 50.7   & 114            \\
         Cascade Mask R-CNN\cite{cai_cascade_2017}                           & Swin-T\cite{liu_swin_2021-1}                                                                                              & IN1k                                                              & 50.4   & 745            \\
         \rowcolor[gray]{0.95}
         TinySaver($t=0.25$)                                                 & \begin{tabular}[l]{@{}l@{}}Co-DINO-Swin-L\cite{zong_detrs_2023-1} \\      + YOLOv8n\cite{yolov8_ultralytics}\end{tabular} & Obj365\cite{shao_objects365_2019}, IN1k                           & 51.5   & 589            \\
         Cascade Mask R-CNN\cite{cai_cascade_2017}                           & Swin-S\cite{liu_swin_2021-1}                                                                                              & IN1k                                                              & 51.9   & 838            \\
         -                                                                   & YOLOv8l\cite{yolov8_ultralytics}                                                                                          & IN1k                                                              & 52.9   & 165            \\
         Mask-RCNN\cite{he_mask_2018}                                        & ConvNeXt V2-B\cite{woo_convnext_2023-1}                                                                                   & IN1k                                                              & 52.9   & 486            \\
         \rowcolor[gray]{0.95}
         TinySaver($t=0.26$)                                                 & \begin{tabular}[l]{@{}l@{}}Co-DINO-Swin-L\cite{zong_detrs_2023-1}\\      +YOLOv8n\cite{yolov8_ultralytics}\end{tabular}   & Obj365\cite{shao_objects365_2019}, IN1k                           & 53.1   & 645            \\
         RT-DETR\cite{lv_detrs_2023}                                         & R50\cite{he_deep_2016}                                                                                                    & IN1k                                                              & 53.1   & 136            \\
         -                                                                   & YOLOv8x\cite{yolov8_ultralytics}                                                                                          & IN1k                                                              & 53.9   & 258            \\
         HTC++\cite{chen_hybrid_2019}                                        & Swin-B\cite{liu_swin_2021-1}                                                                                              & IN22k                                                             & 56.4   & 1043           \\
         \rowcolor[gray]{0.95}
         TinySaver($t=0.34$)                                                 & \begin{tabular}[l]{@{}l@{}}Co-DINO-Swin-L\cite{zong_detrs_2023-1}\\      +YOLOv8x\cite{yolov8_ultralytics}\end{tabular}   & Obj365\cite{shao_objects365_2019}, IN1k                           & 56.5   & 671            \\
         HTC++\cite{chen_hybrid_2019}                                        & Swin-L\cite{liu_swin_2021-1}                                                                                              & IN22k                                                             & 57.1   & 1470           \\
         \rowcolor[gray]{0.95}
         TinySaver($t=0.37$)                                                 & \begin{tabular}[l]{@{}l@{}}Co-DINO-Swin-L\cite{zong_detrs_2023-1}\\      +YOLOv8n\cite{yolov8_ultralytics}\end{tabular}   & Obj365\cite{shao_objects365_2019}, IN1k                           & 60.3   & 1076           \\
         Soft-teacher\cite{xu_end--end_2021-1}, HTC++\cite{chen_hybrid_2019} & Swin-L\cite{liu_swin_2021-1}                                                                                              & Obj365\cite{shao_objects365_2019}                                 & 60.1   & 1470           \\
         \rowcolor[gray]{0.95}
         TinySaver($t=0.69$)                                                 & \begin{tabular}[l]{@{}l@{}}Co-DINO-Swin-L\cite{zong_detrs_2023-1}\\+YOLOv8n\cite{yolov8_ultralytics}\end{tabular}         & Obj365\cite{shao_objects365_2019}, IN1k                           & 63.0   & 1497           \\
         Co-DINO\cite{zong_detrs_2023-1}                                     & Swin-L\cite{liu_swin_2021-1}                                                                                              & Obj365\cite{shao_objects365_2019}                                 & 64.1   & 1527$^\dagger$ \\
         \hline
         \multicolumn{5}{l}{}                                                                                                                                                                                                                                                                          \\
         \multicolumn{5}{l}{$^\dagger$The FLOPs is measured by us, otherwise are collected from corresponding work.}
      \end{tabular}%
   \end{adjustbox}
\end{wraptable}

\subsection{Test on COCO detection}

We extended our experiments to the COCO dataset\cite{lin_microsoft_2015} for object detection, utilizing the scores of detected boxes as the confidence metric. Specifically, we filter out boxes with scores below $0.05$ and use the average score of the remaining boxes as the model's confidence for each sample. We employs a real-time YOLOv8\cite{yolov8_ultralytics} detectors as savers. Images not confidently predicted by YOLOv8\cite{yolov8_ultralytics} are passed to a larger model. Results compared with other detectors are presented in \cref{tab:result-coco}. The proposed method demonstrates superiority in several scenarios. While Co-DINO-Swin-L\cite{zong_detrs_2023-1} excels in high complexity scenarios, YOLOv8\cite{yolov8_ultralytics} offers greater efficiency though with constrained performance. TinySaver manages to extend their advantages to other complexity ranges. However, in object detection, it does not outperform all static models as it does in classification, likely due to bounding box quality being less correlated with score. This outcome aligns with our analysis in \cref{sec:moe-efficiency} and suggests further exploration of external evaluators for such tasks.

\section{Conclusion}

This paper explores the feasibility and effectiveness of using tiny models to decrease the computation of large models. We introduce TinySaver, a novel methodology that employs the most appropriate tiny model to reduce the workload of large models on simpler samples, thus reducing overall complexity. Furthermore, we analyze our approach within early exit (EE) and Mixture of Experts (MoE) frameworks and discuss its broader application in complexity reduction. Our methods have undergone thorough testing with various models which are designed and trained in different ways, showing significant computational reductions across most of them, highlighting the potential of integrating tiny models into extensive AI systems to save on computation.

\\
\section*{Acknowledgment}
This publication has emanated from research conducted with the financial support of 1) Science Foundation Ireland under Grant number 18/CRT/6183 2) Horizon 2020 ERA-NET Chist-Era grant, CHIST-ERA-18-ACAI-003 and 3) Microelectronic Circuits Centre Ireland. For the purpose of Open Access, the author has applied a CC BY public copyright licence to any Author Accepted Manuscript version arising from 
this submission.

\bibliographystyle{splncs04}

\clearpage

\section{Appendix}
\subsection{Full experiment results}
\label{sec:exp-details}

In addition to our main paper examples, we test our method across a diverse range of 44 pre-trained models. These models vary in scale, architecture, development year, and purpose. Our method consistently reduces computation significantly in all cases, with minimal or no loss in accuracy.
\subsubsection{Pool of models}
Models in our experiments include:

\begin{enumerate}
    \item ConvNeXtv2\cite{woo_convnext_2023-1}: $\mathit{atto}$, $\mathit{femto}$, $\mathit{pico}$, $\mathit{nano}$, $\mathit{tiny}$, $\mathit{base}$, $\mathit{large}$, $\mathit{huge}$
    \item DaViT\cite{ding_davit_2022}: $\mathit{tiny}$, $\mathit{small}$, $\mathit{base}$
    \item MaxViT\cite{tu_maxvit_2022}: $\mathit{tiny}$, $\mathit{small}$, $\mathit{base}$, $\mathit{large}$
    \item SwinT\cite{liu_swin_2021-1}: $\mathit{tiny}$, $\mathit{small}$, $\mathit{base}$, $\mathit{large}$
    \item ResNet\cite{he_deep_2016}: $\mathit{18}$, $\mathit{34}$, $\mathit{50}$, $\mathit{101}$, $\mathit{152}$
    \item EfficientNet\cite{tan_efficientnet_2019-1}: $\mathit{b0}$, $\mathit{b1}$, $\mathit{b2}$, $\mathit{b3}$, $\mathit{b4}$
    \item EfficientViT\cite{cai_efficientvit_2023}: $\mathit{b0}$, $\mathit{b1}$, $\mathit{b2}$, $\mathit{b3}$
    \item EfficientViT\cite{liu_efficientvit_2023}: $\mathit{m1}$, $\mathit{m2}$, $\mathit{m3}$, $\mathit{m4}$, $\mathit{m5}$
    \item EfficientFormerv2\cite{li_rethinking_2023}: $\mathit{s0}$, $\mathit{s1}$, $\mathit{s2}$, $\mathit{l}$
    \item MobileNetv3\cite{howard_searching_2019}: $\mathit{small100}$, $\mathit{large100}$
\end{enumerate}
All these models are trained in their original ways with the input size of $224\times 224$ for the ImageNet-1k classification task. Some models are pre-trained with larger datasets\cite{woo_convnext_2023-1}. The model weights of ResNet is provided by torchvision\cite{TorchVision_maintainers_and_contributors_TorchVision_PyTorch_s_Computer_2016}, all other models are provided by timm\cite{wightman_rwightmanpytorch-image-models_2023} and their complexities are measured by the profiling tool of DeepSpeed\cite{rasley_deepspeed_2020}. Each model can employ a smaller one as its saver, as described in \cref{sec:saver-selection}.

All models were trained using their original procedures with an input size of $224\times 224$ for the ImageNet-1k classification task. Some models are pre-trained on larger datasets\cite{woo_convnext_2023-1}. ResNet weights are available from torchvision\cite{TorchVision_maintainers_and_contributors_TorchVision_PyTorch_s_Computer_2016}, while weights for other models are accessed via timm\cite{wightman_rwightmanpytorch-image-models_2023}. We assessed their complexities using DeepSpeed's profiling tool\cite{rasley_deepspeed_2020}. According to the methodology described in \cref{sec:saver-selection}, each model could utilize a smaller one as its saver.
\subsubsection{Compression effect}


\cref{tab:comp-full-2} complements \cref{tab:result-main}, providing extensive experimental results that reinforce our primary findings discussed in \cref{sec:results-main}. Typically, larger models exhibit more substantial compression potential and can maintain their performance even after significant reductions. In contrast, compressing smaller models presents challenges, as their corresponding saver models are relatively larger in proportion. The smallest models, EfficientViT$_\mathit{m0}$ and MobileNetv3$_\mathit{small100}$, are not used as base models due to the absence of smaller eligible savers. Moreover, for effective zero-loss compression, saver models need to closely approximate the accuracy of their base models; otherwise, a trade-off between performance and compression is inevitable.

\begin{table}[]
    \resizebox{\textwidth}{!}{%
        \begin{tabular}{lrr|lrr|llll|ll}
            \hline
            \multicolumn{3}{l|}{Base Model}                             & \multicolumn{3}{l|}{Saver Model} & \multicolumn{4}{c|}{Required MACs(\%) when Acc. Drop $\leq$} & \multicolumn{2}{c}{Max   Performance}

                                                                                                                       \\
            Name                                                        & GFLOPs.                          & Acc.                                                  
       & Name                                                        & GFLOPs. & Acc.     & \multicolumn{1}{r}{1\%}                                                
             & \multicolumn{1}{r}{0.5\%}                                                           & \multicolumn{1}{r}{0.1\%}                                    
                      & \multicolumn{1}{r|}{0\%}                                                            & \multicolumn{1}{r}{Avg. GMACs}                       
                               & \multicolumn{1}{r}{Acc.}                                                              \\ \hline
            \rowcolor[gray]{0.95}
            ConvNeXtV2$_\mathit{huge}$\cite{woo_convnext_2023-1}        & 114.92                           & 86.25\%                                               
      & EfficientFormerV2$_\mathit{l}$\cite{li_rethinking_2023}     & 2.51    & 83.53\%  & \begin{tabular}[c]{@{}l@{}}13.21\\      \textcolor{mygreen}{(-88.5\%)}\end{tabular} & \begin{tabular}[c]{@{}l@{}}19.41\\      \textcolor{mygreen}{(-83.1\%)}\end{tabular} & \begin{tabular}[c]{@{}l@{}}31.17\\      \textcolor{mygreen}{(-72.9\%)}\end{tabular} & \begin{tabular}[c]{@{}l@{}}67.34\\      \textcolor{mygreen}{(-41.4\%)}\end{tabular} & \begin{tabular}[c]{@{}l@{}}70.70\\      \textcolor{mygreen}{(-38.5\%)}\end{tabular} & \begin{tabular}[c]{@{}l@{}}86.25\%\\      \textcolor{mygreen}{(+0.00\%)}\end{tabular} \\
            MaxViT$_\mathit{large}$\cite{tu_maxvit_2022}                & 42.80                            & 84.82\%                                               
      & EfficientFormerV2$_\mathit{l}$\cite{li_rethinking_2023}     & 2.51    & 83.53\%  & \begin{tabular}[c]{@{}l@{}}3.46\\      \textcolor{mygreen}{(-91.9\%)}\end{tabular}  & \begin{tabular}[c]{@{}l@{}}5.48\\      \textcolor{mygreen}{(-87.2\%)}\end{tabular}  & \begin{tabular}[c]{@{}l@{}}7.81\\      \textcolor{mygreen}{(-81.8\%)}\end{tabular}  & \begin{tabular}[c]{@{}l@{}}9.04\\      \textcolor{mygreen}{(-78.9\%)}\end{tabular}  & \begin{tabular}[c]{@{}l@{}}11.24\\      \textcolor{mygreen}{(-73.7\%)}\end{tabular} & \begin{tabular}[c]{@{}l@{}}84.92\%\\      \textcolor{mygreen}{(+0.09\%)}\end{tabular} \\
            \rowcolor[gray]{0.95}
            ConvNeXtV2$_\mathit{large}$\cite{woo_convnext_2023-1}       & 34.36                            & 85.76\%                                               
      & EfficientFormerV2$_\mathit{l}$\cite{li_rethinking_2023}     & 2.51    & 83.53\%  & \begin{tabular}[c]{@{}l@{}}5.16\\      \textcolor{mygreen}{(-85.0\%)}\end{tabular}  & \begin{tabular}[c]{@{}l@{}}7.15\\      \textcolor{mygreen}{(-79.2\%)}\end{tabular}  & \begin{tabular}[c]{@{}l@{}}11.00\\      \textcolor{mygreen}{(-68.0\%)}\end{tabular} & \begin{tabular}[c]{@{}l@{}}15.15\\      \textcolor{mygreen}{(-55.9\%)}\end{tabular} & \begin{tabular}[c]{@{}l@{}}17.49\\      \textcolor{mygreen}{(-49.1\%)}\end{tabular} & \begin{tabular}[c]{@{}l@{}}85.78\%\\      \textcolor{mygreen}{(+0.02\%)}\end{tabular} \\
            Swin$_\mathit{large}$\cite{liu_swin_2021-1}                 & 34.02                            & 86.24\%                                               
      & EfficientFormerV2$_\mathit{l}$\cite{li_rethinking_2023}     & 2.51    & 83.53\%  & \begin{tabular}[c]{@{}l@{}}5.56\\      \textcolor{mygreen}{(-83.7\%)}\end{tabular}  & \begin{tabular}[c]{@{}l@{}}7.22\\      \textcolor{mygreen}{(-78.8\%)}\end{tabular}  & \begin{tabular}[c]{@{}l@{}}10.95\\      \textcolor{mygreen}{(-67.8\%)}\end{tabular} & \begin{tabular}[c]{@{}l@{}}14.46\\      \textcolor{mygreen}{(-57.5\%)}\end{tabular} & \begin{tabular}[c]{@{}l@{}}18.70\\      \textcolor{mygreen}{(-45.0\%)}\end{tabular} & \begin{tabular}[c]{@{}l@{}}86.26\%\\      \textcolor{mygreen}{(+0.02\%)}\end{tabular} \\
            \rowcolor[gray]{0.95}
            MaxViT$_\mathit{base}$\cite{tu_maxvit_2022}                 & 23.39                            & 84.80\%                                               
      & EfficientFormerV2$_\mathit{l}$\cite{li_rethinking_2023}     & 2.51    & 83.53\%  & \begin{tabular}[c]{@{}l@{}}2.95\\      \textcolor{mygreen}{(-87.4\%)}\end{tabular}  & \begin{tabular}[c]{@{}l@{}}4.13\\      \textcolor{mygreen}{(-82.3\%)}\end{tabular}  & \begin{tabular}[c]{@{}l@{}}5.33\\      \textcolor{mygreen}{(-77.2\%)}\end{tabular}  & \begin{tabular}[c]{@{}l@{}}6.74\\      \textcolor{mygreen}{(-71.2\%)}\end{tabular}  & \begin{tabular}[c]{@{}l@{}}7.28\\      \textcolor{mygreen}{(-68.9\%)}\end{tabular}  & \begin{tabular}[c]{@{}l@{}}84.86\%\\      \textcolor{mygreen}{(+0.06\%)}\end{tabular} \\
            ConvNeXtV2$_\mathit{base}$\cite{woo_convnext_2023-1}        & 15.35                            & 84.87\%                                               
      & EfficientFormerV2$_\mathit{s2}$\cite{li_rethinking_2023}    & 1.22    & 82.15\% & \begin{tabular}[c]{@{}l@{}}2.74\\      \textcolor{mygreen}{(-82.1\%)}\end{tabular}  & \begin{tabular}[c]{@{}l@{}}3.57\\      \textcolor{mygreen}{(-76.8\%)}\end{tabular}  & \begin{tabular}[c]{@{}l@{}}4.85\\      \textcolor{mygreen}{(-68.4\%)}\end{tabular}  & \begin{tabular}[c]{@{}l@{}}5.44\\      \textcolor{mygreen}{(-64.6\%)}\end{tabular}  & \begin{tabular}[c]{@{}l@{}}6.35\\      \textcolor{mygreen}{(-58.6\%)}\end{tabular}  & \begin{tabular}[c]{@{}l@{}}84.92\%\\      \textcolor{mygreen}{(+0.05\%)}\end{tabular} \\
            \rowcolor[gray]{0.95}
            DaViT$_\mathit{base}$\cite{ding_davit_2022}                 & 15.22                            & 84.48\%                                               
      & EfficientFormerV2$_\mathit{s2}$\cite{li_rethinking_2023}    & 1.22    & 82.15\% & \begin{tabular}[c]{@{}l@{}}2.40\\      \textcolor{mygreen}{(-84.3\%)}\end{tabular}  & \begin{tabular}[c]{@{}l@{}}3.16\\      \textcolor{mygreen}{(-79.2\%)}\end{tabular}  & \begin{tabular}[c]{@{}l@{}}4.40\\      \textcolor{mygreen}{(-71.1\%)}\end{tabular}  & \begin{tabular}[c]{@{}l@{}}5.19\\      \textcolor{mygreen}{(-65.9\%)}\end{tabular}  & \begin{tabular}[c]{@{}l@{}}6.50\\      \textcolor{mygreen}{(-57.3\%)}\end{tabular}  & \begin{tabular}[c]{@{}l@{}}84.53\%\\      \textcolor{mygreen}{(+0.04\%)}\end{tabular} \\
            Swin$_\mathit{base}$\cite{liu_swin_2021-1}                  & 15.13                            & 85.14\%                                               
      & EfficientFormerV2$_\mathit{s2}$\cite{li_rethinking_2023}    & 1.22    & 82.15\% & \begin{tabular}[c]{@{}l@{}}2.81\\      \textcolor{mygreen}{(-81.4\%)}\end{tabular}  & \begin{tabular}[c]{@{}l@{}}3.70\\      \textcolor{mygreen}{(-75.6\%)}\end{tabular}  & \begin{tabular}[c]{@{}l@{}}4.79\\      \textcolor{mygreen}{(-68.3\%)}\end{tabular}  & \begin{tabular}[c]{@{}l@{}}9.06\\      \textcolor{mygreen}{(-40.1\%)}\end{tabular}  & \begin{tabular}[c]{@{}l@{}}9.40\\      \textcolor{mygreen}{(-37.9\%)}\end{tabular}  & \begin{tabular}[c]{@{}l@{}}85.15\%\\      \textcolor{mygreen}{(+0.00\%)}\end{tabular} \\
            \rowcolor[gray]{0.95}
            ResNet$_\mathit{152}$\cite{he_deep_2016}                    & 11.51                            & 78.24\%                                               
      & EfficientViT$_\mathit{m2}$\cite{liu_efficientvit_2023}      & 0.20    & 70.78\% & \begin{tabular}[c]{@{}l@{}}4.17\\      \textcolor{mygreen}{(-63.8\%)}\end{tabular}  & \begin{tabular}[c]{@{}l@{}}4.91\\      \textcolor{mygreen}{(-57.3\%)}\end{tabular}  & \begin{tabular}[c]{@{}l@{}}6.30\\      \textcolor{mygreen}{(-45.3\%)}\end{tabular}  & \begin{tabular}[c]{@{}l@{}}8.79\\      \textcolor{mygreen}{(-23.7\%)}\end{tabular}  & \begin{tabular}[c]{@{}l@{}}9.92\\      \textcolor{mygreen}{(-13.8\%)}\end{tabular}  & \begin{tabular}[c]{@{}l@{}}78.27\%\\      \textcolor{mygreen}{(+0.03\%)}\end{tabular} \\
            MaxViT$_\mathit{small}$\cite{tu_maxvit_2022}                & 11.31                            & 84.33\%                                               
      & EfficientFormerV2$_\mathit{s2}$\cite{li_rethinking_2023}    & 1.22    & 82.15\% & \begin{tabular}[c]{@{}l@{}}2.01\\      \textcolor{mygreen}{(-82.2\%)}\end{tabular}  & \begin{tabular}[c]{@{}l@{}}2.51\\      \textcolor{mygreen}{(-77.8\%)}\end{tabular}  & \begin{tabular}[c]{@{}l@{}}3.53\\      \textcolor{mygreen}{(-68.8\%)}\end{tabular}  & \begin{tabular}[c]{@{}l@{}}3.86\\      \textcolor{mygreen}{(-65.9\%)}\end{tabular}  & \begin{tabular}[c]{@{}l@{}}4.29\\      \textcolor{mygreen}{(-62.1\%)}\end{tabular}  & \begin{tabular}[c]{@{}l@{}}84.42\%\\      \textcolor{mygreen}{(+0.08\%)}\end{tabular} \\
            \rowcolor[gray]{0.95}
            DaViT$_\mathit{small}$\cite{ding_davit_2022}                & 8.58                             & 84.00\%                                               
      & EfficientFormerV2$_\mathit{s2}$\cite{li_rethinking_2023}    & 1.22    & 82.15\% & \begin{tabular}[c]{@{}l@{}}1.65\\      \textcolor{mygreen}{(-80.8\%)}\end{tabular}  & \begin{tabular}[c]{@{}l@{}}2.06\\      \textcolor{mygreen}{(-76.0\%)}\end{tabular}  & \begin{tabular}[c]{@{}l@{}}2.92\\      \textcolor{mygreen}{(-66.0\%)}\end{tabular}  & \begin{tabular}[c]{@{}l@{}}3.25\\      \textcolor{mygreen}{(-62.1\%)}\end{tabular}  & \begin{tabular}[c]{@{}l@{}}3.54\\      \textcolor{mygreen}{(-58.7\%)}\end{tabular}  & \begin{tabular}[c]{@{}l@{}}84.06\%\\      \textcolor{mygreen}{(+0.05\%)}\end{tabular} \\
            Swin$_\mathit{small}$\cite{liu_swin_2021-1}                 & 8.51                             & 83.25\%                                               
       & EfficientNet$_\mathit{b3}$\cite{tan_efficientnet_2019-1}    & 0.96    & 78.63\% & \begin{tabular}[c]{@{}l@{}}2.58\\      \textcolor{mygreen}{(-69.6\%)}\end{tabular}  & \begin{tabular}[c]{@{}l@{}}3.11\\      \textcolor{mygreen}{(-63.5\%)}\end{tabular}  & \begin{tabular}[c]{@{}l@{}}4.09\\      \textcolor{mygreen}{(-51.9\%)}\end{tabular}  & \begin{tabular}[c]{@{}l@{}}4.41\\      \textcolor{mygreen}{(-48.2\%)}\end{tabular}  & \begin{tabular}[c]{@{}l@{}}5.67\\      \textcolor{mygreen}{(-33.4\%)}\end{tabular}  & \begin{tabular}[c]{@{}l@{}}83.34\%\\      \textcolor{mygreen}{(+0.09\%)}\end{tabular} \\
            \rowcolor[gray]{0.95}
            ResNet$_\mathit{101}$\cite{he_deep_2016}                    & 7.80                             & 77.26\%                                               
      & EfficientViT$_\mathit{b0}$\cite{cai_efficientvit_2023}      & 0.10    & 71.35\% & \begin{tabular}[c]{@{}l@{}}2.10\\      \textcolor{mygreen}{(-73.1\%)}\end{tabular}  & \begin{tabular}[c]{@{}l@{}}2.53\\      \textcolor{mygreen}{(-67.5\%)}\end{tabular}  & \begin{tabular}[c]{@{}l@{}}2.93\\      \textcolor{mygreen}{(-62.5\%)}\end{tabular}  & \begin{tabular}[c]{@{}l@{}}3.46\\      \textcolor{mygreen}{(-55.7\%)}\end{tabular}  & \begin{tabular}[c]{@{}l@{}}4.43\\      \textcolor{mygreen}{(-43.3\%)}\end{tabular}  & \begin{tabular}[c]{@{}l@{}}77.38\%\\      \textcolor{mygreen}{(+0.12\%)}\end{tabular} \\
            MaxViT$_\mathit{tiny}$\cite{tu_maxvit_2022}                 & 5.37                             & 83.36\%                                               
       & EfficientFormerV2$_\mathit{s2}$\cite{li_rethinking_2023}    & 1.22    & 82.15\% & \begin{tabular}[c]{@{}l@{}}1.29\\      \textcolor{mygreen}{(-76.0\%)}\end{tabular}  & \begin{tabular}[c]{@{}l@{}}1.52\\      \textcolor{mygreen}{(-71.7\%)}\end{tabular}  & \begin{tabular}[c]{@{}l@{}}1.82\\      \textcolor{mygreen}{(-66.0\%)}\end{tabular}  & \begin{tabular}[c]{@{}l@{}}1.96\\      \textcolor{mygreen}{(-63.5\%)}\end{tabular}  & \begin{tabular}[c]{@{}l@{}}2.57\\      \textcolor{mygreen}{(-52.2\%)}\end{tabular}  & \begin{tabular}[c]{@{}l@{}}83.58\%\\      \textcolor{mygreen}{(+0.22\%)}\end{tabular} \\
            \rowcolor[gray]{0.95}
            ConvNeXtV2$_\mathit{tiny}$\cite{woo_convnext_2023-1}        & 4.46                             & 82.94\%                                               
      & EfficientNet$_\mathit{b2}$\cite{tan_efficientnet_2019-1}    & 0.66    & 77.89\% & \begin{tabular}[c]{@{}l@{}}1.65\\      \textcolor{mygreen}{(-62.9\%)}\end{tabular}  & \begin{tabular}[c]{@{}l@{}}1.98\\      \textcolor{mygreen}{(-55.6\%)}\end{tabular}  & \begin{tabular}[c]{@{}l@{}}2.54\\      \textcolor{mygreen}{(-42.9\%)}\end{tabular}  & \begin{tabular}[c]{@{}l@{}}3.50\\      \textcolor{mygreen}{(-21.4\%)}\end{tabular}  & \begin{tabular}[c]{@{}l@{}}4.54\\      \textcolor{myred}{(+2.0\%)}\end{tabular}     & \begin{tabular}[c]{@{}l@{}}82.96\%\\      \textcolor{mygreen}{(+0.01\%)}\end{tabular} \\
            DaViT$_\mathit{tiny}$\cite{ding_davit_2022}                 & 4.41                             & 82.71\%                                               
      & EfficientFormerV2$_\mathit{s2}$\cite{li_rethinking_2023}    & 1.22    & 82.15\% & -                                                                        
           & \begin{tabular}[c]{@{}l@{}}1.23\\      \textcolor{mygreen}{(-72.1\%)}\end{tabular}  & \begin{tabular}[c]{@{}l@{}}1.37\\      \textcolor{mygreen}{(-68.9\%)}\end{tabular}  & \begin{tabular}[c]{@{}l@{}}1.42\\      \textcolor{mygreen}{(-67.9\%)}\end{tabular}  & \begin{tabular}[c]{@{}l@{}}2.06\\      \textcolor{mygreen}{(-53.3\%)}\end{tabular}  & \begin{tabular}[c]{@{}l@{}}83.23\%\\      \textcolor{mygreen}{(+0.51\%)}\end{tabular} \\
            \rowcolor[gray]{0.95}
            Swin$_\mathit{tiny}$\cite{liu_swin_2021-1}                  & 4.35                             & 80.89\%                                               
      & EfficientNet$_\mathit{b0}$\cite{tan_efficientnet_2019-1}    & 0.39    & 77.70\% & \begin{tabular}[c]{@{}l@{}}0.91\\      \textcolor{mygreen}{(-79.0\%)}\end{tabular}  & \begin{tabular}[c]{@{}l@{}}1.15\\      \textcolor{mygreen}{(-73.5\%)}\end{tabular}  & \begin{tabular}[c]{@{}l@{}}1.42\\      \textcolor{mygreen}{(-67.4\%)}\end{tabular}  & \begin{tabular}[c]{@{}l@{}}1.54\\      \textcolor{mygreen}{(-64.5\%)}\end{tabular}  & \begin{tabular}[c]{@{}l@{}}2.07\\      \textcolor{mygreen}{(-52.3\%)}\end{tabular}  & \begin{tabular}[c]{@{}l@{}}81.09\%\\      \textcolor{mygreen}{(+0.19\%)}\end{tabular} \\
            ResNet$_\mathit{50}$\cite{he_deep_2016}                     & 4.09                             & 75.85\%                                               
       & EfficientViT$_\mathit{b0}$\cite{cai_efficientvit_2023}      & 0.10    & 71.35\% & \begin{tabular}[c]{@{}l@{}}0.93\\      \textcolor{mygreen}{(-77.2\%)}\end{tabular}  & \begin{tabular}[c]{@{}l@{}}1.11\\      \textcolor{mygreen}{(-72.9\%)}\end{tabular}  & \begin{tabular}[c]{@{}l@{}}1.31\\      \textcolor{mygreen}{(-67.9\%)}\end{tabular}  & \begin{tabular}[c]{@{}l@{}}1.36\\      \textcolor{mygreen}{(-66.7\%)}\end{tabular}  & \begin{tabular}[c]{@{}l@{}}1.87\\      \textcolor{mygreen}{(-54.2\%)}\end{tabular}  & \begin{tabular}[c]{@{}l@{}}76.15\%\\      \textcolor{mygreen}{(+0.30\%)}\end{tabular} \\
            \rowcolor[gray]{0.95}
            EfficientViT$_\mathit{b3}$\cite{cai_efficientvit_2023}      & 3.88                             & 83.14\%                                               
      & EfficientNet$_\mathit{b0}$\cite{tan_efficientnet_2019-1}    & 0.39    & 77.70\% & \begin{tabular}[c]{@{}l@{}}1.33\\      \textcolor{mygreen}{(-65.6\%)}\end{tabular}  & \begin{tabular}[c]{@{}l@{}}1.59\\      \textcolor{mygreen}{(-59.0\%)}\end{tabular}  & \begin{tabular}[c]{@{}l@{}}2.08\\      \textcolor{mygreen}{(-46.4\%)}\end{tabular}  & \begin{tabular}[c]{@{}l@{}}3.17\\      \textcolor{mygreen}{(-18.4\%)}\end{tabular}  & \begin{tabular}[c]{@{}l@{}}3.89\\      \textcolor{myred}{(+0.3\%)}\end{tabular}     & \begin{tabular}[c]{@{}l@{}}83.16\%\\      \textcolor{mygreen}{(+0.01\%)}\end{tabular} \\
            ResNet$_\mathit{34}$\cite{he_deep_2016}                     & 3.66                             & 73.2\%                                                
       & MobileNetV3$_\mathit{small100}$\cite{howard_searching_2019} & 0.06    & 67.65\% & \begin{tabular}[c]{@{}l@{}}1.11\\      \textcolor{mygreen}{(-69.7\%)}\end{tabular}  & \begin{tabular}[c]{@{}l@{}}1.35\\      \textcolor{mygreen}{(-63.0\%)}\end{tabular}  & \begin{tabular}[c]{@{}l@{}}1.57\\      \textcolor{mygreen}{(-57.3\%)}\end{tabular}  & \begin{tabular}[c]{@{}l@{}}1.63\\      \textcolor{mygreen}{(-55.6\%)}\end{tabular}  & \begin{tabular}[c]{@{}l@{}}2.01\\      \textcolor{mygreen}{(-45.1\%)}\end{tabular}  & \begin{tabular}[c]{@{}l@{}}73.36\%\\      \textcolor{mygreen}{(+0.16\%)}\end{tabular} \\
            \rowcolor[gray]{0.95}
            EfficientFormerV2$_\mathit{l}$\cite{li_rethinking_2023}     & 2.51                             & 83.53\%                                               
       & MobileNetV3$_\mathit{large100}$\cite{howard_searching_2019} & 0.22    & 75.78\%  & \begin{tabular}[c]{@{}l@{}}0.99\\      \textcolor{mygreen}{(-60.6\%)}\end{tabular}  & \begin{tabular}[c]{@{}l@{}}1.17\\      \textcolor{mygreen}{(-53.4\%)}\end{tabular}  & \begin{tabular}[c]{@{}l@{}}1.49\\      \textcolor{mygreen}{(-40.6\%)}\end{tabular}  & \begin{tabular}[c]{@{}l@{}}2.51\\      \textcolor{myred}{(+0.1\%)}\end{tabular}     & \begin{tabular}[c]{@{}l@{}}2.51\\      \textcolor{myred}{(+0.1\%)}\end{tabular}     & \begin{tabular}[c]{@{}l@{}}83.53\%\\      \textcolor{mygreen}{(+0.00\%)}\end{tabular} \\
            ConvNeXtV2$_\mathit{nano}$\cite{woo_convnext_2023-1}        & 2.45                             & 81.87\%                                               
       & MobileNetV3$_\mathit{large100}$\cite{howard_searching_2019} & 0.22    & 75.78\%  & \begin{tabular}[c]{@{}l@{}}0.88\\      \textcolor{mygreen}{(-63.9\%)}\end{tabular}  & \begin{tabular}[c]{@{}l@{}}1.04\\      \textcolor{mygreen}{(-57.5\%)}\end{tabular}  & \begin{tabular}[c]{@{}l@{}}1.34\\      \textcolor{mygreen}{(-45.2\%)}\end{tabular}  & \begin{tabular}[c]{@{}l@{}}1.76\\      \textcolor{mygreen}{(-28.2\%)}\end{tabular}  & \begin{tabular}[c]{@{}l@{}}2.45\\      \textcolor{myred}{(+0.3\%)}\end{tabular}     & \begin{tabular}[c]{@{}l@{}}81.87\%\\      \textcolor{mygreen}{(+0.00\%)}\end{tabular} \\
            \rowcolor[gray]{0.95}
            ResNet$_\mathit{18}$\cite{he_deep_2016}                     & 1.81                             & 69.53\%                                               
      & MobileNetV3$_\mathit{small100}$\cite{howard_searching_2019} & 0.06    & 67.65\% & \begin{tabular}[c]{@{}l@{}}0.19\\      \textcolor{mygreen}{(-89.4\%)}\end{tabular}  & \begin{tabular}[c]{@{}l@{}}0.28\\      \textcolor{mygreen}{(-84.3\%)}\end{tabular}  & \begin{tabular}[c]{@{}l@{}}0.37\\      \textcolor{mygreen}{(-79.5\%)}\end{tabular}  & \begin{tabular}[c]{@{}l@{}}0.41\\      \textcolor{mygreen}{(-77.2\%)}\end{tabular}  & \begin{tabular}[c]{@{}l@{}}0.84\\      \textcolor{mygreen}{(-54.0\%)}\end{tabular}  & \begin{tabular}[c]{@{}l@{}}70.31\%\\      \textcolor{mygreen}{(+0.78\%)}\end{tabular} \\
            EfficientViT$_\mathit{b2}$\cite{cai_efficientvit_2023}      & 1.55                             & 81.91\%                                               
       & MobileNetV3$_\mathit{large100}$\cite{howard_searching_2019} & 0.22    & 75.78\%  & \begin{tabular}[c]{@{}l@{}}0.64\\      \textcolor{mygreen}{(-58.9\%)}\end{tabular}  & \begin{tabular}[c]{@{}l@{}}0.74\\      \textcolor{mygreen}{(-52.5\%)}\end{tabular}  & \begin{tabular}[c]{@{}l@{}}0.95\\      \textcolor{mygreen}{(-38.6\%)}\end{tabular}  & \begin{tabular}[c]{@{}l@{}}1.65\\      \textcolor{myred}{(+6.2\%)}\end{tabular}     & \begin{tabular}[c]{@{}l@{}}1.65\\      \textcolor{myred}{(+6.2\%)}\end{tabular}     & \begin{tabular}[c]{@{}l@{}}81.91\%\\      \textcolor{mygreen}{(+0.00\%)}\end{tabular} \\
            \rowcolor[gray]{0.95}
            EfficientNet$_\mathit{b4}$\cite{tan_efficientnet_2019-1}    & 1.50                             & 79.40\%                                               
      & MobileNetV3$_\mathit{large100}$\cite{howard_searching_2019} & 0.22    & 75.78\%  & \begin{tabular}[c]{@{}l@{}}0.46\\      \textcolor{mygreen}{(-69.4\%)}\end{tabular}  & \begin{tabular}[c]{@{}l@{}}0.54\\      \textcolor{mygreen}{(-63.8\%)}\end{tabular}  & \begin{tabular}[c]{@{}l@{}}0.64\\      \textcolor{mygreen}{(-57.7\%)}\end{tabular}  & \begin{tabular}[c]{@{}l@{}}0.66\\      \textcolor{mygreen}{(-56.1\%)}\end{tabular}  & \begin{tabular}[c]{@{}l@{}}0.78\\      \textcolor{mygreen}{(-48.4\%)}\end{tabular}  & \begin{tabular}[c]{@{}l@{}}79.66\%\\      \textcolor{mygreen}{(+0.26\%)}\end{tabular} \\
            ConvNeXtV2$_\mathit{pico}$\cite{woo_convnext_2023-1}        & 1.37                             & 80.31\%                                               
       & MobileNetV3$_\mathit{large100}$\cite{howard_searching_2019} & 0.22    & 75.78\%  & \begin{tabular}[c]{@{}l@{}}0.52\\      \textcolor{mygreen}{(-61.9\%)}\end{tabular}  & \begin{tabular}[c]{@{}l@{}}0.61\\      \textcolor{mygreen}{(-55.4\%)}\end{tabular}  & \begin{tabular}[c]{@{}l@{}}0.71\\      \textcolor{mygreen}{(-48.0\%)}\end{tabular}  & \begin{tabular}[c]{@{}l@{}}0.81\\      \textcolor{mygreen}{(-40.6\%)}\end{tabular}  & \begin{tabular}[c]{@{}l@{}}1.05\\      \textcolor{mygreen}{(-23.1\%)}\end{tabular}  & \begin{tabular}[c]{@{}l@{}}80.38\%\\      \textcolor{mygreen}{(+0.07\%)}\end{tabular} \\
            \rowcolor[gray]{0.95}
            EfficientFormerV2$_\mathit{s2}$\cite{li_rethinking_2023}    & 1.22                             & 82.15\%                                               
      & MobileNetV3$_\mathit{large100}$\cite{howard_searching_2019} & 0.22    & 75.78\%  & \begin{tabular}[c]{@{}l@{}}0.57\\      \textcolor{mygreen}{(-53.2\%)}\end{tabular}  & \begin{tabular}[c]{@{}l@{}}0.64\\      \textcolor{mygreen}{(-47.2\%)}\end{tabular}  & \begin{tabular}[c]{@{}l@{}}0.82\\      \textcolor{mygreen}{(-33.0\%)}\end{tabular}  & \begin{tabular}[c]{@{}l@{}}1.02\\      \textcolor{mygreen}{(-16.2\%)}\end{tabular}  & \begin{tabular}[c]{@{}l@{}}1.05\\      \textcolor{mygreen}{(-13.8\%)}\end{tabular}  & \begin{tabular}[c]{@{}l@{}}82.17\%\\      \textcolor{mygreen}{(+0.01\%)}\end{tabular} \\
            EfficientNet$_\mathit{b3}$\cite{tan_efficientnet_2019-1}    & 0.96                             & 78.63\%                                               
      & MobileNetV3$_\mathit{large100}$\cite{howard_searching_2019} & 0.22    & 75.78\%  & \begin{tabular}[c]{@{}l@{}}0.33\\      \textcolor{mygreen}{(-66.0\%)}\end{tabular}  & \begin{tabular}[c]{@{}l@{}}0.38\\      \textcolor{mygreen}{(-61.0\%)}\end{tabular}  & \begin{tabular}[c]{@{}l@{}}0.43\\      \textcolor{mygreen}{(-55.4\%)}\end{tabular}  & \begin{tabular}[c]{@{}l@{}}0.45\\      \textcolor{mygreen}{(-53.2\%)}\end{tabular}  & \begin{tabular}[c]{@{}l@{}}0.55\\      \textcolor{mygreen}{(-42.4\%)}\end{tabular}  & \begin{tabular}[c]{@{}l@{}}79.01\%\\      \textcolor{mygreen}{(+0.38\%)}\end{tabular} \\
            \rowcolor[gray]{0.95}
            ConvNeXtV2$_\mathit{femto}$\cite{woo_convnext_2023-1}       & 0.78                             & 78.49\%                                               
      & MobileNetV3$_\mathit{large100}$\cite{howard_searching_2019} & 0.22    & 75.78\%  & \begin{tabular}[c]{@{}l@{}}0.31\\      \textcolor{mygreen}{(-60.2\%)}\end{tabular}  & \begin{tabular}[c]{@{}l@{}}0.36\\      \textcolor{mygreen}{(-53.4\%)}\end{tabular}  & \begin{tabular}[c]{@{}l@{}}0.42\\      \textcolor{mygreen}{(-46.7\%)}\end{tabular}  & \begin{tabular}[c]{@{}l@{}}0.43\\      \textcolor{mygreen}{(-44.6\%)}\end{tabular}  & \begin{tabular}[c]{@{}l@{}}0.51\\      \textcolor{mygreen}{(-34.9\%)}\end{tabular}  & \begin{tabular}[c]{@{}l@{}}78.76\%\\      \textcolor{mygreen}{(+0.27\%)}\end{tabular} \\
            EfficientNet$_\mathit{b2}$\cite{tan_efficientnet_2019-1}    & 0.66                             & 77.89\%                                               
      & MobileNetV3$_\mathit{large100}$\cite{howard_searching_2019} & 0.22    & 75.78\%  & \begin{tabular}[c]{@{}l@{}}0.27\\      \textcolor{mygreen}{(-59.4\%)}\end{tabular}  & \begin{tabular}[c]{@{}l@{}}0.30\\      \textcolor{mygreen}{(-53.9\%)}\end{tabular}  & \begin{tabular}[c]{@{}l@{}}0.34\\      \textcolor{mygreen}{(-47.8\%)}\end{tabular}  & \begin{tabular}[c]{@{}l@{}}0.36\\      \textcolor{mygreen}{(-45.5\%)}\end{tabular}  & \begin{tabular}[c]{@{}l@{}}0.46\\      \textcolor{mygreen}{(-30.9\%)}\end{tabular}  & \begin{tabular}[c]{@{}l@{}}78.29\%\\      \textcolor{mygreen}{(+0.39\%)}\end{tabular} \\
            \rowcolor[gray]{0.95}
            EfficientFormerV2$_\mathit{s1}$\cite{li_rethinking_2023}    & 0.63                             & 79.69\%                                               
      & MobileNetV3$_\mathit{large100}$\cite{howard_searching_2019} & 0.22    & 75.78\%  & \begin{tabular}[c]{@{}l@{}}0.34\\      \textcolor{mygreen}{(-46.7\%)}\end{tabular}  & \begin{tabular}[c]{@{}l@{}}0.38\\      \textcolor{mygreen}{(-39.9\%)}\end{tabular}  & \begin{tabular}[c]{@{}l@{}}0.43\\      \textcolor{mygreen}{(-31.4\%)}\end{tabular}  & \begin{tabular}[c]{@{}l@{}}0.50\\      \textcolor{mygreen}{(-21.4\%)}\end{tabular}  & \begin{tabular}[c]{@{}l@{}}0.58\\      \textcolor{mygreen}{(-8.6\%)}\end{tabular}   & \begin{tabular}[c]{@{}l@{}}79.76\%\\      \textcolor{mygreen}{(+0.06\%)}\end{tabular} \\
            EfficientNet$_\mathit{b1}$\cite{tan_efficientnet_2019-1}    & 0.57                             & 77.57\%                                               
      & EfficientViT$_\mathit{b0}$\cite{cai_efficientvit_2023}      & 0.10    & 71.35\% & \begin{tabular}[c]{@{}l@{}}0.26\\      \textcolor{mygreen}{(-54.9\%)}\end{tabular}  & \begin{tabular}[c]{@{}l@{}}0.28\\      \textcolor{mygreen}{(-50.2\%)}\end{tabular}  & \begin{tabular}[c]{@{}l@{}}0.33\\      \textcolor{mygreen}{(-42.4\%)}\end{tabular}  & \begin{tabular}[c]{@{}l@{}}0.35\\      \textcolor{mygreen}{(-38.8\%)}\end{tabular}  & \begin{tabular}[c]{@{}l@{}}0.38\\      \textcolor{mygreen}{(-33.3\%)}\end{tabular}  & \begin{tabular}[c]{@{}l@{}}77.64\%\\      \textcolor{mygreen}{(+0.06\%)}\end{tabular} \\
            \rowcolor[gray]{0.95}
            ConvNeXtV2$_\mathit{atto}$\cite{woo_convnext_2023-1}        & 0.55                             & 76.64\%                                               
      & MobileNetV3$_\mathit{large100}$\cite{howard_searching_2019} & 0.22    & 75.78\%  & -                                                                       
            & \begin{tabular}[c]{@{}l@{}}0.23\\      \textcolor{mygreen}{(-57.6\%)}\end{tabular}  & \begin{tabular}[c]{@{}l@{}}0.26\\      \textcolor{mygreen}{(-52.8\%)}\end{tabular}  & \begin{tabular}[c]{@{}l@{}}0.26\\      \textcolor{mygreen}{(-51.7\%)}\end{tabular}  & \begin{tabular}[c]{@{}l@{}}0.39\\      \textcolor{mygreen}{(-29.6\%)}\end{tabular}  & \begin{tabular}[c]{@{}l@{}}77.33\%\\      \textcolor{mygreen}{(+0.68\%)}\end{tabular} \\
            EfficientViT$_\mathit{m5}$\cite{liu_efficientvit_2023}      & 0.52                             & 77.08\%                                               
      & MobileNetV3$_\mathit{large100}$\cite{howard_searching_2019} & 0.22    & 75.78\%  & \begin{tabular}[c]{@{}l@{}}0.23\\      \textcolor{mygreen}{(-55.7\%)}\end{tabular}  & \begin{tabular}[c]{@{}l@{}}0.25\\      \textcolor{mygreen}{(-51.2\%)}\end{tabular}  & \begin{tabular}[c]{@{}l@{}}0.28\\      \textcolor{mygreen}{(-44.9\%)}\end{tabular}  & \begin{tabular}[c]{@{}l@{}}0.29\\      \textcolor{mygreen}{(-43.0\%)}\end{tabular}  & \begin{tabular}[c]{@{}l@{}}0.38\\      \textcolor{mygreen}{(-27.1\%)}\end{tabular}  & \begin{tabular}[c]{@{}l@{}}77.55\%\\      \textcolor{mygreen}{(+0.47\%)}\end{tabular} \\
            \rowcolor[gray]{0.95}
            EfficientViT$_\mathit{b1}$\cite{cai_efficientvit_2023}      & 0.51                             & 79.12\%                                               
       & MobileNetV3$_\mathit{large100}$\cite{howard_searching_2019} & 0.22    & 75.78\%  & \begin{tabular}[c]{@{}l@{}}0.30\\      \textcolor{mygreen}{(-40.8\%)}\end{tabular}  & \begin{tabular}[c]{@{}l@{}}0.34\\      \textcolor{mygreen}{(-33.4\%)}\end{tabular}  & \begin{tabular}[c]{@{}l@{}}0.37\\      \textcolor{mygreen}{(-26.8\%)}\end{tabular}  & \begin{tabular}[c]{@{}l@{}}0.40\\      \textcolor{mygreen}{(-22.6\%)}\end{tabular}  & \begin{tabular}[c]{@{}l@{}}0.45\\      \textcolor{mygreen}{(-11.5\%)}\end{tabular}  & \begin{tabular}[c]{@{}l@{}}79.20\%\\      \textcolor{mygreen}{(+0.08\%)}\end{tabular} \\
            EfficientNet$_\mathit{b0}$\cite{tan_efficientnet_2019-1}    & 0.39                             & 77.70\%                                               
      & EfficientViT$_\mathit{b0}$\cite{cai_efficientvit_2023}      & 0.10    & 71.35\% & \begin{tabular}[c]{@{}l@{}}0.21\\      \textcolor{mygreen}{(-45.6\%)}\end{tabular}  & \begin{tabular}[c]{@{}l@{}}0.23\\      \textcolor{mygreen}{(-39.5\%)}\end{tabular}  & \begin{tabular}[c]{@{}l@{}}0.29\\      \textcolor{mygreen}{(-25.2\%)}\end{tabular}  & \begin{tabular}[c]{@{}l@{}}0.33\\      \textcolor{mygreen}{(-14.6\%)}\end{tabular}  & \begin{tabular}[c]{@{}l@{}}0.36\\      \textcolor{mygreen}{(-7.2\%)}\end{tabular}   & \begin{tabular}[c]{@{}l@{}}77.71\%\\      \textcolor{mygreen}{(+0.01\%)}\end{tabular} \\
            \rowcolor[gray]{0.95}
            EfficientFormerV2$_\mathit{s0}$\cite{li_rethinking_2023}    & 0.38                             & 76.24\%                                               
      & EfficientViT$_\mathit{b0}$\cite{cai_efficientvit_2023}      & 0.10    & 71.35\% & \begin{tabular}[c]{@{}l@{}}0.18\\      \textcolor{mygreen}{(-51.0\%)}\end{tabular}  & \begin{tabular}[c]{@{}l@{}}0.20\\      \textcolor{mygreen}{(-46.0\%)}\end{tabular}  & \begin{tabular}[c]{@{}l@{}}0.22\\      \textcolor{mygreen}{(-41.3\%)}\end{tabular}  & \begin{tabular}[c]{@{}l@{}}0.23\\      \textcolor{mygreen}{(-38.8\%)}\end{tabular}  & \begin{tabular}[c]{@{}l@{}}0.26\\      \textcolor{mygreen}{(-30.6\%)}\end{tabular}  & \begin{tabular}[c]{@{}l@{}}76.38\%\\      \textcolor{mygreen}{(+0.13\%)}\end{tabular} \\
            EfficientViT$_\mathit{m4}$\cite{liu_efficientvit_2023}      & 0.30                             & 74.32\%                                               
      & MobileNetV3$_\mathit{small100}$\cite{howard_searching_2019} & 0.06    & 67.65\% & \begin{tabular}[c]{@{}l@{}}0.16\\      \textcolor{mygreen}{(-47.3\%)}\end{tabular}  & \begin{tabular}[c]{@{}l@{}}0.18\\      \textcolor{mygreen}{(-40.3\%)}\end{tabular}  & \begin{tabular}[c]{@{}l@{}}0.21\\      \textcolor{mygreen}{(-29.2\%)}\end{tabular}  & \begin{tabular}[c]{@{}l@{}}0.23\\      \textcolor{mygreen}{(-21.6\%)}\end{tabular}  & \begin{tabular}[c]{@{}l@{}}0.26\\      \textcolor{mygreen}{(-12.9\%)}\end{tabular}  & \begin{tabular}[c]{@{}l@{}}74.34\%\\      \textcolor{mygreen}{(+0.02\%)}\end{tabular} \\
            \rowcolor[gray]{0.95}
            EfficientViT$_\mathit{m3}$\cite{liu_efficientvit_2023}      & 0.26                             & 73.38\%                                               
      & MobileNetV3$_\mathit{small100}$\cite{howard_searching_2019} & 0.06    & 67.65\% & \begin{tabular}[c]{@{}l@{}}0.14\\      \textcolor{mygreen}{(-46.2\%)}\end{tabular}  & \begin{tabular}[c]{@{}l@{}}0.16\\      \textcolor{mygreen}{(-39.7\%)}\end{tabular}  & \begin{tabular}[c]{@{}l@{}}0.18\\      \textcolor{mygreen}{(-30.8\%)}\end{tabular}  & \begin{tabular}[c]{@{}l@{}}0.19\\      \textcolor{mygreen}{(-26.6\%)}\end{tabular}  & \begin{tabular}[c]{@{}l@{}}0.21\\      \textcolor{mygreen}{(-18.3\%)}\end{tabular}  & \begin{tabular}[c]{@{}l@{}}73.45\%\\      \textcolor{mygreen}{(+0.06\%)}\end{tabular} \\
            MobileNetV3$_\mathit{large100}$\cite{howard_searching_2019} & 0.22                             & 75.78\%                                               
       & MobileNetV3$_\mathit{small100}$\cite{howard_searching_2019} & 0.06    & 67.65\% & \begin{tabular}[c]{@{}l@{}}0.14\\      \textcolor{mygreen}{(-33.6\%)}\end{tabular}  & \begin{tabular}[c]{@{}l@{}}0.16\\      \textcolor{mygreen}{(-26.5\%)}\end{tabular}  & \begin{tabular}[c]{@{}l@{}}0.19\\      \textcolor{mygreen}{(-12.2\%)}\end{tabular}  & \begin{tabular}[c]{@{}l@{}}0.22\\      \textcolor{myred}{(+2.8\%)}\end{tabular}     & \begin{tabular}[c]{@{}l@{}}0.22\\      \textcolor{myred}{(+3.4\%)}\end{tabular}     & \begin{tabular}[c]{@{}l@{}}75.78\%\\      \textcolor{mygreen}{(+0.00\%)}\end{tabular} \\
            \rowcolor[gray]{0.95}
            EfficientViT$_\mathit{m2}$\cite{liu_efficientvit_2023}      & 0.20                             & 70.78\%                                               
      & MobileNetV3$_\mathit{small100}$\cite{howard_searching_2019} & 0.06    & 67.65\% & \begin{tabular}[c]{@{}l@{}}0.09\\      \textcolor{mygreen}{(-54.5\%)}\end{tabular}  & \begin{tabular}[c]{@{}l@{}}0.10\\      \textcolor{mygreen}{(-48.3\%)}\end{tabular}  & \begin{tabular}[c]{@{}l@{}}0.12\\      \textcolor{mygreen}{(-40.9\%)}\end{tabular}  & \begin{tabular}[c]{@{}l@{}}0.12\\      \textcolor{mygreen}{(-38.8\%)}\end{tabular}  & \begin{tabular}[c]{@{}l@{}}0.15\\      \textcolor{mygreen}{(-22.6\%)}\end{tabular}  & \begin{tabular}[c]{@{}l@{}}71.13\%\\      \textcolor{mygreen}{(+0.35\%)}\end{tabular} \\
            EfficientViT$_\mathit{m1}$\cite{liu_efficientvit_2023}      & 0.16                             & 68.32\%                                               
       & MobileNetV3$_\mathit{small100}$\cite{howard_searching_2019} & 0.06    & 67.65\% & -                                                                       
            & \begin{tabular}[c]{@{}l@{}}0.06\\      \textcolor{mygreen}{(-64.1\%)}\end{tabular}  & \begin{tabular}[c]{@{}l@{}}0.06\\      \textcolor{mygreen}{(-60.7\%)}\end{tabular}  & \begin{tabular}[c]{@{}l@{}}0.07\\      \textcolor{mygreen}{(-59.6\%)}\end{tabular}  & \begin{tabular}[c]{@{}l@{}}0.11\\      \textcolor{mygreen}{(-31.9\%)}\end{tabular}  & \begin{tabular}[c]{@{}l@{}}69.41\%\\      \textcolor{mygreen}{(+1.09\%)}\end{tabular} \\
            \rowcolor[gray]{0.95}
            EfficientViT$_\mathit{b0}$\cite{cai_efficientvit_2023}      & 0.10                             & 71.35\%                                               
      & MobileNetV3$_\mathit{small100}$\cite{howard_searching_2019} & 0.06    & 67.65\% & \begin{tabular}[c]{@{}l@{}}0.08\\      \textcolor{mygreen}{(-16.5\%)}\end{tabular}  & \begin{tabular}[c]{@{}l@{}}0.09\\      \textcolor{mygreen}{(-8.7\%)}\end{tabular}   & \begin{tabular}[c]{@{}l@{}}0.10\\      \textcolor{mygreen}{(-0.6\%)}\end{tabular}   & \begin{tabular}[c]{@{}l@{}}0.10\\      \textcolor{myred}{(+3.7\%)}\end{tabular}     & \begin{tabular}[c]{@{}l@{}}0.11\\      \textcolor{myred}{(+11.5\%)}\end{tabular}    & \begin{tabular}[c]{@{}l@{}}71.46\%\\      \textcolor{mygreen}{(+0.10\%)}\end{tabular} \\
            \hline
        \end{tabular}%
    }

    \caption{Compression effect while TinySaver applied to all models in the pool}
    \label{tab:comp-full-2}
\end{table}
\begin{figure}
    \centering
    \includegraphics[width=\linewidth]{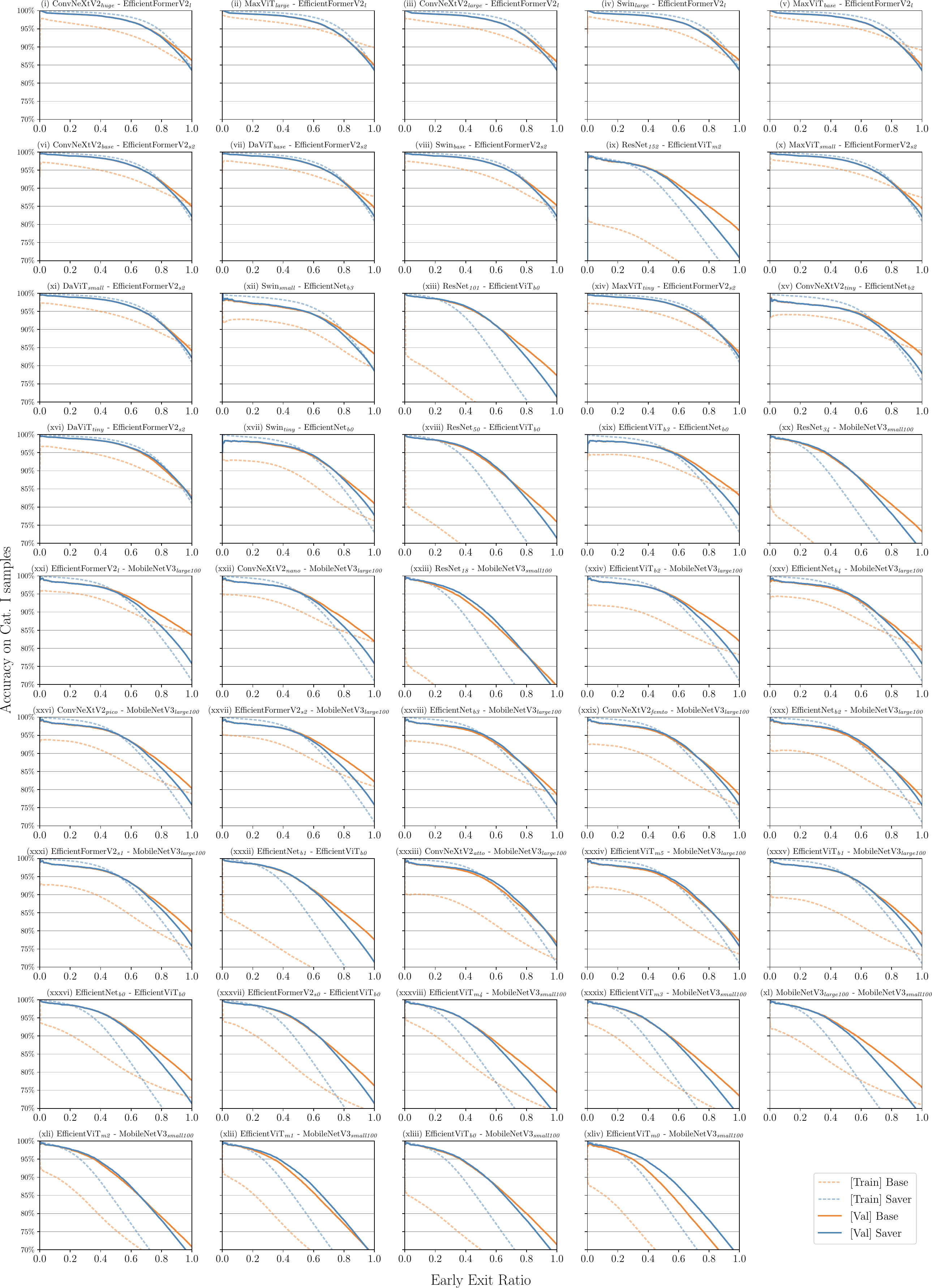}
    \caption{\textbf{Accuracy vs early exit ratio} on the ImageNet-1k training/validation set. The intersection of solid lines denotes the ratio where the saver and base model have the equivalent accuracy on the validation set. The intersection of dashed lines are for the training set. Every plot is labelled as \textbf{Base - Saver}.}
    \label{fig:match-performance-full}
\end{figure}
\subsubsection{Performance and early exit ratio}

\cref{fig:match-performance-full} expands on \cref{fig:match-performance}, illustrating how accuracy varies as more samples are processed by the saver model. In all test cases, the saver model initially excels on the training set. Cat. I samples are filtered based on saver's confidence and the saver can perform exceptionally well with high-confidence samples. While base models struggle with these same samples, leading to poorer performance when Cat. I is small. Additionally, data augmentation complicates early performance for base models. However, as Cat. I expands to include more samples, the situation changes. Finally, if we set saver's threshold as 0 and let all samples be Cat. I, the base will apparently outperform the saver.

On the validation set, the significant initial advantage of the saver model decreases, but it remains more accurate until a later intersection point. This consistent trend across all plots demonstrates the reliability of models when they are highly confident, which is the source of compression effect.

Nonetheless, the base model's underperformance on the training set results in a relative overconfidence in the saver model. Although the validation set is not directly impacted, this discrepancy could adversely affect the accuracy of saver selection, particularly in multi-exit systems like ESN, which rely on exit policies derived from training data. Introducing more data augmentation on the training set may help reduce the discrepancy between training and validation performances.
\subsection{Speed test}
In our experiments, we use original pre-trained models, which are implemented differently by each author. Their heterogeneity makes them unevenly supported on different platforms. Therefore, it's difficult to compare FLOPs reduction and actual speedups directly. However, for a given deployment scenario, we can replace FLOPs with measured latency while identifying the best saver model. The speed test on a PC with i9-11900F and one RTX 3090 is below. We ran inference on IN-1k val set, and recorded the model throughput. All TinySaver enabled models were configured to have no accuracy loss. \cref{tab:speed-test} shows the results. Though models exhibit varing efficiency on different deployment scenarios, we can still observe that TinySaver can significantly improve the throughput of the base model.
\begin{table}[h]
    \resizebox{\linewidth}{!}{%
        \begin{tabular}{c|cccc}
            \hline
            \multirow{3}{*}{Base Model}                      & \multicolumn{4}{c}{{ Saver}}                                                                                                                                                                                \\
                                                             & \multicolumn{4}{c}{Original$\to$Ours   Avg. throughput img/s}                                                                                                                                               \\ \cline{2-5}
                                                             & \multicolumn{1}{c|}{{ bs=1,CPU}}                              & \multicolumn{1}{c|}{{ bs=1}}                    & \multicolumn{1}{c|}{{ bs=128}}                     & {{ bs=512}}                          \\ \hline
            \multirow{2}{*}{ConvNextv2$_\mathit{h}$}         & \multicolumn{1}{c|}{{ Swin$_\mathit{b}$}}                     & \multicolumn{1}{c|}{{ ConvNextv2$_\mathit{l}$}} & \multicolumn{1}{c|}{{ Swin$_\mathit{b}$}}          & \multirow{2}{*}{OOM} \\
                                                             & \multicolumn{1}{c|}{1.9$\to$6.4}                              & \multicolumn{1}{c|}{55.0$\to$98.5}              & \multicolumn{1}{c|}{96.9$\to$307.8}                &                                      \\ \hline
            \multirow{2}{*}{Swin$_\mathit{b}$}               & \multicolumn{1}{c|}{{ EfficientFormerV2$_\mathit{l}$}}        & \multicolumn{1}{c|}{{ ConvNextv2$_\mathit{b}$}} & \multicolumn{1}{c|}{{ DaViT$_\mathit{t}$}}         & { DaViT$_\mathit{t}$}                \\
                                                             & \multicolumn{1}{c|}{9.9$\to$17.7}                             & \multicolumn{1}{c|}{97.2$\to$117.5}             & \multicolumn{1}{c|}{487.0$\to$659.7}               & 496.3$\to$688.8                      \\ \hline
            \multirow{2}{*}{EfficientNet$_\mathit{b2}$}      & \multicolumn{1}{c|}{{ ConvNextv2$_\mathit{a}$}}               & \multicolumn{1}{c|}{{ ConvNextv2$_\mathit{a}$}} & \multicolumn{1}{c|}{{ EfficientViT$_\mathit{m5}$}} & { EfficientViT$_\mathit{m5}$}        \\
                                                             & \multicolumn{1}{c|}{51.9$\to$81.4}                            & \multicolumn{1}{c|}{150.9$\to$229.4}            & \multicolumn{1}{c|}{2172.1$\to$2649.7}             & 2237.5$\to$3344.4                    \\ \hline
            \multirow{2}{*}{EfficientFormerV2$_\mathit{s2}$} & \multicolumn{1}{c|}{{ ConvNextv2$_\mathit{n}$}}               & \multicolumn{1}{c|}{{ ConvNextv2$_\mathit{n}$}} & \multicolumn{1}{c|}{{ EfficientViT$_\mathit{b2}$}} & { EfficientViT$_\mathit{b2}$}        \\
                                                             & \multicolumn{1}{c|}{34.0$\to$42.0}                            & \multicolumn{1}{c|}{83.8$\to$222.3}             & \multicolumn{1}{c|}{562.1$\to$1339.9}              & 570.8$\to$1542.8                     \\ \hline
        \end{tabular}%
    }
\footnotetext{The RTX 3090 24GB is out-of-memory for this case}
\caption{Average throughput of models with TinySaver enabled on different deployment scenarios}
\label{tab:speed-test}

\end{table}
\subsection{Exit Sequence Network (ESN)}

\begin{figure}[h]
    \centering
    \includegraphics[width=1\linewidth]{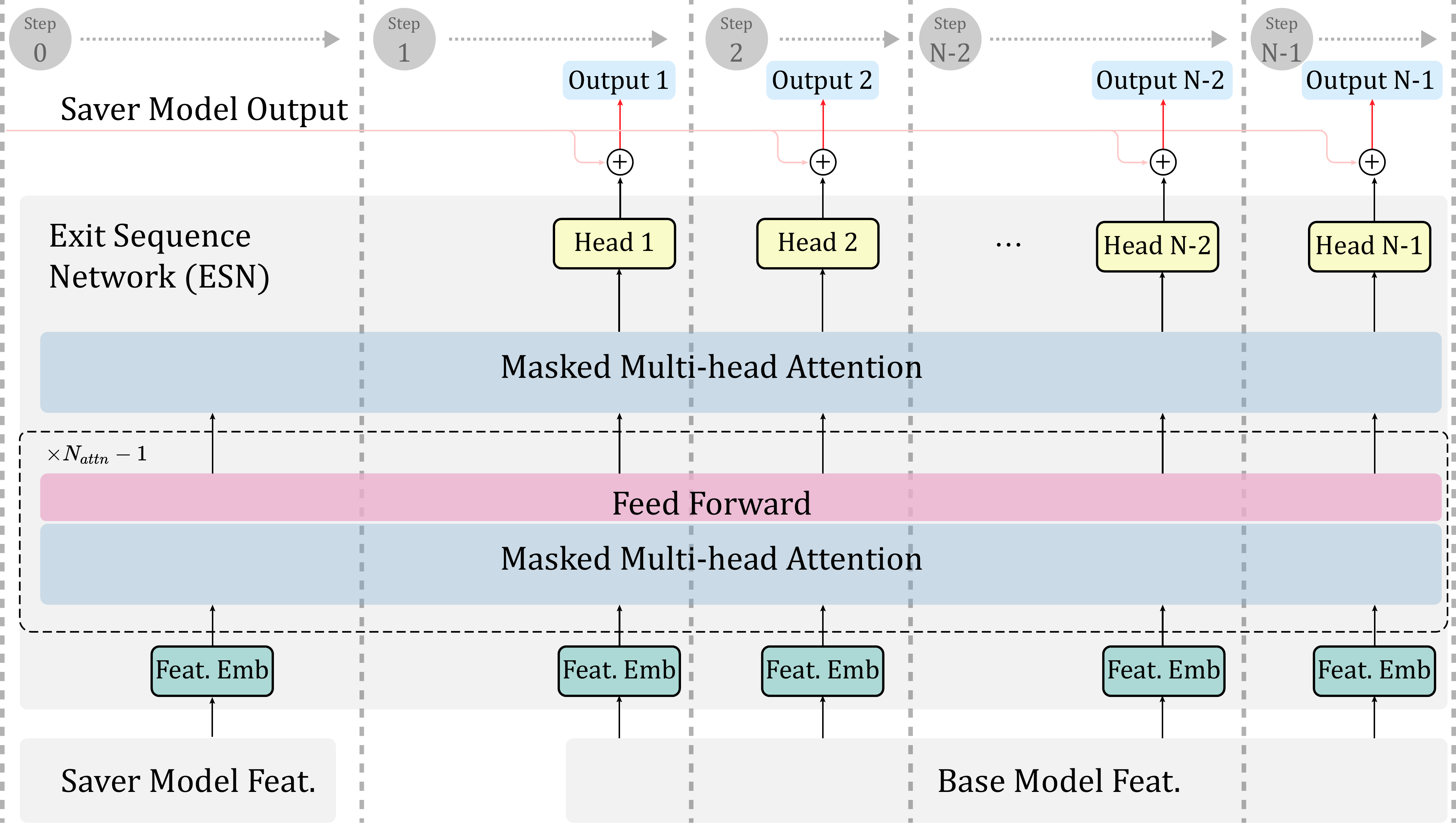}
    \caption{Detailed ESN}
    \label{fig:esn-detail}
\end{figure}
\label{sec:esn_detail}

The design of the ESN comprises three components: (1) Embedding units that transform features at various scales into a format compatible with the subsequent sequential model. (2) A sequential feature bus designed to aggregate features from all preceding steps and transmit them to subsequent stages. (3) Head units, responsible for outputting results like regular attached exits.

To ensure efficiency, the overall scale of the ESN should be kept moderate, thereby minimizing the introduction of overhead. While over-simplified networks might not be able to keep its functions. Careful tuning should be required for the best practice of ESN. We illustrate the detail of ESN in \cref{fig:esn-detail}

\subsubsection{Feature embedding}
In our approach, both saver and base models are considered well-trained and remain unchanged during ESN training. Their features, treated as a sequence of tokens, need embedding prior to joining the attention layer. In our ESN design, we employ small networks each containing a global average pooling layer and a linear layer, to embed intermediate backbone features.

\subsubsection{Sequential feature bus}
The primary objective of this part is to ensure the delivery of the most informative feature for each head unit. The sequential execution is mandatory, aligning with the step-by-step nature of the dynamic inference process. For this purpose, we have chosen to implement a transformer decoder network recognized for its effectiveness across a broad of applications. Meanwhile, the first step, i.e. features from the saver model, are critical in our system. The transformer allows for direct referencing of these features without requiring intermediary states. This is another reason to select transformers. Contrasting with the commonly used Feature Pyramid Network (FPN)\cite{lin_feature_2017-1}, our Exit Sequence Network (ESN) does not accept backward connections from deeper layers. This is required by EE to inference progressively in practice. Therefore, our implementation uses mask to maintain causality.

\subsubsection{Exiting heads}

Exiting heads are responsible for generating the output. We also brings the saver model's output, making $\tilde{\boldsymbol{y}}_{\mathit{n}}=\tilde{\boldsymbol{y}}_{\mathit{S}}+\Delta \tilde{\boldsymbol{y}}_n$. The weight of ESN heads are initialized as zero and only need to learn $\Delta \tilde{\boldsymbol{y}}_n$. Therefore, the intermediate outputs combine the information from the saver and base model and their performance is lower bounded by the saver model.

\subsection{Configuration and training details of ESN}
\label{sec:esn_training_detail}


\cref{fig:esn-detail} shows detailed structure of our prototype design of the ESN. We extract intermediate features from each residual block's output, integrating them with saver features into the ESN of multi-head causal attention layers. Subsequently, single-layer classifiers at each exit generate predictions, all supervised by the loss function detailed in \cref{sec:loss-function}. \cref{tab:esn-param} includes the hyper-parameter of training ESN. We run a basic grid search of suitable attention layer numbers and dims.

\cref{fig:esn-detail} illustrates the detailed structure of our ESN prototype. We extract intermediate features from the output of each backbone segment and combine them with features from the saver into the ESN, which includes multi-head causal attention layers. Then, single-layer classifiers at each exit are responsible for generating predictions, trained by the loss function detailed in \cref{sec:loss-function}. Additionally, \cref{tab:esn-param} lists the hyperparameters used for training the ESN, where we conducted a basic grid search to identify optimal numbers of attention layers and their dimensions.

\begin{table}[h]
    \centering
    \resizebox{0.3\linewidth}{!}{
        \begin{tabular}{l|l}
            config                 & value                                    \\
            \specialrule{.1em}{.05em}{.05em}
            optimizer              & AdamW                                    \\
            learning rate          & $6.25 \times 10^{-4}$                    \\
            learning rate schedule & cosine decay                             \\
            warmup epochs          & 20                                       \\
            augmentation           & AutoAugment\cite{cubuk_autoaugment_2019} \\
            mixup                  & 0.8                                      \\
            cutmix                 & 1.0                                      \\
            training epochs        & 10                                       \\
            num attn heads         & 8                                        \\
            num attn layers        & \{0, 1, 2, 3\}                           \\
            dim                    & \{8, 32, 128, 512\}                      \\
        \end{tabular}}
    \caption{Configurations for training ESN}
    \label{tab:esn-param}
\end{table}
\subsubsection{Loss function}
\label{sec:loss-function}

In traditional Early Exit (EE) models, the training usually employs the original loss function or knowledge distillation. However, in our approach, a significant portion of inputs is processed solely by the saver model, meaning intermediate exits do not influence these particular inputs. Additionally, lightweight networks have limited capacity for knowledge representation, suggesting the utility of reducing the burden on such networks from samples unlikely to reach intermediate exits. Consequently, we adjust the weight of samples in the loss function based on the saver model's confidence levels. The loss function applied at the $n^{th}$ exit is outlined in \cref{eq:loss}, reflecting this adaptation to reduce the load on intermediate exits while maintaining effective learning.

\begin{align}
    L_n= \sum_{m=0}^{M} 2(1- \mathop{max}_{i}(\tilde{\boldsymbol{y}}^{(m,i)}_{\mathit{S}}))\mathcal{L}(\tilde{\boldsymbol{y}}^{(m)}_{\mathit{n}})  \label{eq:loss}
\end{align}
Where $\mathop{max}_{i}(\tilde{\boldsymbol{y}}^{(m,i)}_{\mathit{S}})$ is the confidence of the saver model. $\mathcal{L}(\tilde{\boldsymbol{y}}^{(m)}_{\mathit{n}})$ is the regular loss function per sample and we use cross entropy in the experiment of this paper.

\end{document}